%% file: main.tex
\pdfoutput=1
\documentclass[11pt]{article}

\usepackage[final]{acl}

\usepackage[T1]{fontenc}
\usepackage[utf8]{inputenc}

\usepackage{microtype}

\usepackage{inconsolata}

\usepackage{graphicx}

\usepackage{times}
\usepackage{latexsym}

\usepackage{adjustbox}
\usepackage{multirow}
\usepackage{booktabs}
\usepackage{arydshln}

\usepackage{enumitem}
\usepackage{amsmath,amssymb,amsfonts} 

\usepackage{algorithm}
\usepackage[noend]{algpseudocode}

\newtheorem{theorem}{Theorem}[section]

\usepackage[capitalize]{cleveref}

\definecolor{commentcolor}{RGB}{110,154,155}   %

\setlist[itemize]{itemsep=0pt, parsep=0pt}

\definecolor{mygray}{RGB}{224,224,224}
\usepackage{colortbl}

\definecolor{lightblue}{RGB}{100,224,224}

\title{GradOT: Training-free Gradient-preserving Offsite-tuning \\ for Large Language Models}

\author{
 \textbf{Kai Yao\textsuperscript{\rm 1,\rm 2}\thanks{These authors contributed equally to this work.}},
 \textbf{Zhaorui Tan\textsuperscript{\rm 3}\footnotemark[1]},
 \textbf{Penglei Gao\textsuperscript{\rm 4}},
 \textbf{Lichun Li\textsuperscript{\rm 2}},
 \textbf{Kaixin Wu\textsuperscript{\rm 2}},
 \\
 \textbf{Yinggui Wang\textsuperscript{\rm 2}},
 \textbf{Yuan Zhao\textsuperscript{\rm 2}},
 \textbf{Yixin Ji\textsuperscript{\rm 5}},
 \textbf{Wei Wang\textsuperscript{\rm 2}\thanks{Corresponding authors.}},
 \textbf{Jianke Zhu\textsuperscript{\rm 1}\footnotemark[2]},
\\
 \textsuperscript{1}Zhejiang University \ 
 \textsuperscript{2}Ant Group \ 
 \textsuperscript{3}University of Liverpool \ \\
 \textsuperscript{4}Cleveland Clinic Lerner Research Institution \
  \textsuperscript{5}Soochow University
\\
\href{mailto:jiumo.yk@antgroup.com}{jiumo.yk@antgroup.com}, \href{mailto:raytan@liverpool.ac.uk}{raytan@liverpool.ac.uk}
}

\begin{document}
\maketitle

\renewcommand{\thefootnote}{\fnsymbol{footnote}}
\begin{abstract}
The rapid growth of large language models (LLMs) with traditional centralized fine-tuning emerges as a key technique for adapting these models to domain-specific challenges, yielding privacy risks for both model and data owners. One promising solution, called offsite-tuning (OT), is proposed to address these challenges, where a weaker emulator is compressed from the original model and further fine-tuned with adapter to enhance privacy. However, the existing OT-based methods require high computational costs and lack theoretical analysis. This paper introduces a novel OT approach based on gradient-preserving compression, named GradOT. By analyzing the OT problem through the lens of optimization, we propose a method that selectively applies compression techniques such as rank compression and channel pruning, preserving the gradients of fine-tuned adapters while ensuring privacy. Extensive experiments demonstrate that our approach surpasses existing OT methods, both in terms of privacy protection and model performance. Our method provides a theoretical foundation for OT and offers a practical, training-free solution for offsite-tuning of large-scale LLMs.
\end{abstract}

\section{Introduction}
Recent years have witnessed the rapid growth of large language models (LLMs)\cite{bert,gpt2,LLaMA,du2021glm,dong2022survey, wies2024learnability}, whose centralized fine-tuning~\cite{wortsman2022robust,zhou2022conditional, wei2021finetuned,ouyang2022training,tans,liu2024visual,yao-etal-2024-layer} is a common approach for adapting them to more complex, domain-specific tasks. However, the required co-location of the model and data raises risks of jeopardizing the privacy of both their owners~\cite{chua2023fedpeat, OT,li2020bert,zou2023universal,bdatk,liu2024lora}, posing significant barriers to LLMs' applications on sensitive downstream fine-tuning. 

\begin{figure}[t]
\centering
\includegraphics[width=0.9\linewidth]{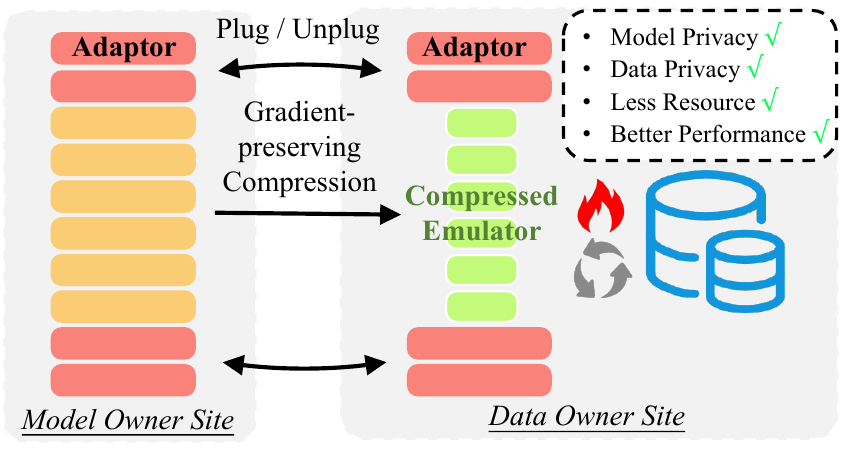}
\caption{Illustration of our proposed Gradient-preserving Offsite-tuning (GradOT). 
}
\label{fig:banner}
\end{figure}

Offsite-tuning (OT)~\cite{OT} has emerged as a promising solution to safeguard the privacy of both data and model owners. \cref{fig:banner} illustrates the process of our proposed method, which follows the general process of OT methods. As seen from this process diagram, OT-based methods usually involve lossy compression of the large language model into a smaller, weaker version, defined as an emulator on the model owner's side. The adapter is further fine-tuned on the compressed emulator with the data sent by data owners and will be returned to the model owner after fine-tuning, replacing the original adapters to form the fully adapted model. 

There are two crucial objectives in OT-based fine-tuning: (1) \textit{Privacy protecting} aims to intentionally increase the performance gap between the finetuned emulator and the fully finetuned model. (2) \textit{Performance preserving} aims to, on the contrary, improve the performance of the final plug-in model. 
The development of a feasible lossy compression method that optimizes these two objectives is the key challenge in this field. OT~\cite{OT} attempted to achieve offsite-tuning by leveraging LayerDrop~\cite{layerdrop} and knowledge distillation~\cite{sanh2019distilbert,hinton2015distilling}, which requires high computational cost and hinders its application in practices on large-scale LLMs. Following this, CRaSh~\cite{crash} used repetitive shared layers to replace dropped layers, proposing a novel training-free OT method. In addition, another noteworthy method termed ScaleOT~\cite{scaleot} proposed to estimate the layer-wise importance with reinforcement learning and replace the less important layers with lightweight modules, significantly improving the privacy protection and final fine-tuned model performance. These OT-based approaches not only ensure privacy for both the data and the model but also encourage the use of the fully adapted model, fostering a mutually beneficial relationship between the data and model owners. However, they heavily rely on empirical validation and lack theoretical analysis, which limits the development of OT-based methods and their practical applications. 

\begin{figure}[t]
\centering
\includegraphics[width=1.\linewidth]{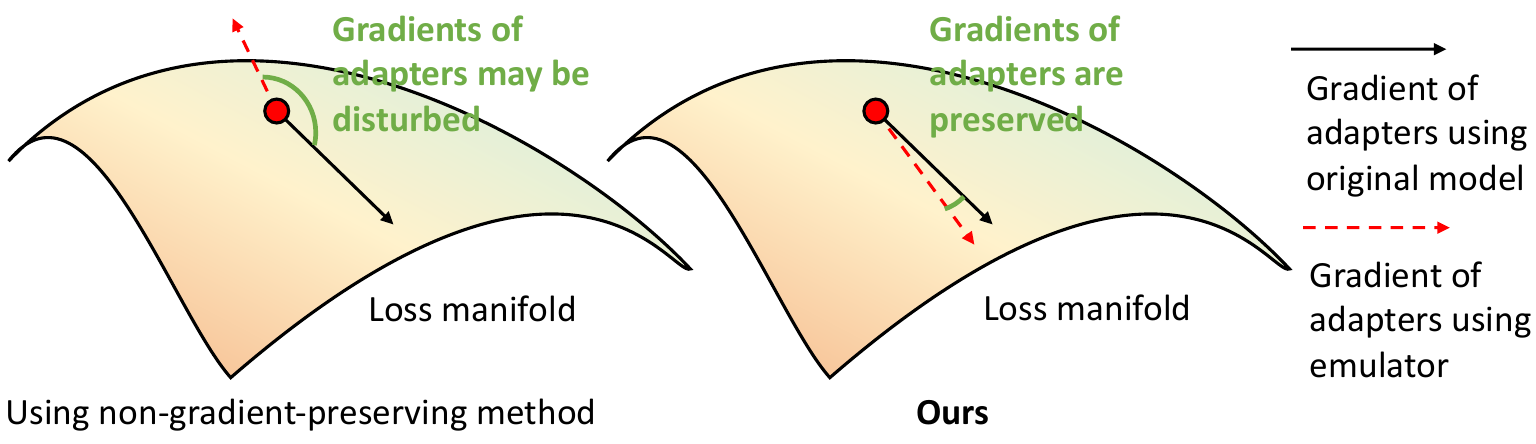}
\caption{Diagram of gradient preserving of adapters through emulators. 
}
\label{fig:gradient_preserve_diagram}
\end{figure}

In this paper, we take the first step in analyzing the OT problem from an optimization perspective. We formally define the OT objective, showing that the problem centers on maximizing privacy while minimizing the gradient discrepancies between the adapters and the compressed emulator, which has not yet been explored in this field. By approximating this formal objective, we introduce a gradient-preserving compression score, which is used to construct the emulator. As shown in \cref{fig:gradient_preserve_diagram}, our gradient-preserving compression score (GSC)-based framework coordinates adapter optimization through joint training with the original model and compressed emulators, achieving dual objectives of maintaining task performance in the final plug-in model and enforcing privacy preservation via gradient consistency mechanisms. As shown in \cref{fig:gradient_preserve_diagram}, our gradient-preserving compression score (GSC)-based method maximizes the adapters that are trained with the original model and emulators, ensuring the performance of the final plug-in model. Then, we propose a novel training-free approach called gradient-preserving offsite tuning (GradOT) as shown in \cref{fig:banner}, which focuses on preserving the gradients of adapters to enhance privacy in the context of OT. Inspired by \citet{scaleot}, our method applies different compression techniques to different blocks within the LLM. Specifically, we use Dynamic Rank Decomposition (DRD) for the MHA and Selective Channel Pruning (SCP) for the MLP. Importantly, our compression is performed based on the scores, ensuring gradient preservation while allowing for increased loss, which enhances both performance and privacy. 

Extensive experiments demonstrate that our method outperforms previous approaches. Our contributions are as follows:
\begin{itemize}
    \item We formally define the OT problem and present its tractable objective. Based on the objectives, we derive the gradient-preserving compression score for OT tasks.
    \item Integrating the gradient-preserving compression score,  we propose a novel training-free OT method named GradOT, which consists of Dynamic Rank Decomposition (DRD) and Selective Channel Pruning (SCP).
    \item Extensive experiments on various LLMs and datasets showcase the effectiveness of our GradOT, validating the efficacy of our gradient-preserving compression score. 
\end{itemize}

\section{Related Work}
\textbf{Large Language Models.}
As models grow in size and complexity, pre-trained Large Language Models (LLMs) have demonstrated impressive performance across various natural language processing (NLP) tasks~\cite{LLaMA,gpt3}. These models possess extensive general knowledge, enabling them to effectively address unknown problems through zero-shot learning or in-context learning~\cite{li2020bert,zou2023universal}.
Nonetheless, when dealing with complex problems, transfer learning with a small dataset remains the preferred choice, as it avoids costly retraining and leverages pretrained knowledge for better performance. Offsite tuning, a subset of transfer learning, aims to utilize the knowledge of LLMs for downstream tasks while ensuring bidirectional privacy between the model and the data sites.

\textbf{Offsite-tuning.}
In contrast to traditional privacy approaches that emphasize data, such as federated learning~\cite{nguyen2021federated} and differential privacy~\cite{dptuning}, offsite tuning methods~\cite{dpopt,OT} prioritize model privacy. Previous research has primarily relied on empirical experiments. For example, OT~\cite{OT} determined through experiments that LayerDrop meets offsite tuning objectives, whereas pruning and quantization are challenging to implement. Building on this, CRaSh~\cite{crash} enhanced performance by sharing remaining layers after layer dropping. ScaleOT~\cite{scaleot} introduced layerwise importance to preserve critical layers and replace less significant ones with lightweight modules. Despite their efficacy, the absence of theoretical analysis has constrained further development. Furthermore, analyzing layer-based methods is challenging due to significant structural differences among various models. This paper offers a theoretical analysis from the perspective of weights and provides a universally effective indicator tailored for offsite tuning.

\section{Method}
\label{sec:method}

\subsection{Theoretical Motivation}
We address privacy concerns that hinder the sharing and co-location of data and LLMs between their respective owners, aiming to tune the model using the data owner's data while avoiding access to the model owner's weights. 
The goal of offsite-tuning is to address the privacy concerns of both data and model owners by indirectly sharing original data and models. Typically, the model owner offers the adapters with an emulator that is compressed from the original model middleweights for data owners to fine-tune on their side. Those tuned adapters would be returned and combined with the original model's middleweights as the final tuned model. 

Following previous works~\cite{scaleot}, the goal of offsite tuning is to find a lossy compression function $\mathcal{F}$ that compresses a full model $\mathcal{M}$ into an alternative, smaller, and weaker model $\widehat{\mathcal{M}}$. Similar to previous works~\cite{OT, crash} and for simplification, we denote ${\mathcal{M}} =\mathcal{A}_2\circ\mathcal{E}\circ\mathcal{A}_1 (\cdot)$, where $\mathcal{A}=[\mathcal{A}_1$, $\mathcal{A}_2]$ denotes the trainable top and bottom layers of full model, and $\mathcal{E}$ represents the frozen middle layers. The compressed version $\widehat{\mathcal{M}} = \widehat{\mathcal{A}}_2\circ\widehat{\mathcal{E}}\circ\widehat{\mathcal{A}}_1 (\cdot)$ where $\widehat{\mathcal{E}} = \mathcal{F}(\mathcal{E}), \widehat{\mathcal{A}}_2 = {\mathcal{A}}_2, \widehat{\mathcal{A}}_1 = {\mathcal{A}}_1$ is used as the initialization for the fine-tuning at the data owner side. Then, the data owner fine-tunes the adapters with $\widehat{\mathcal{E}}$ resulting in the \textit{Emulator Fine-tuned} model denoted as $\widehat{\mathcal{M}}'=\widehat{\mathcal{A}}'_2\circ\widehat{\mathcal{E}}\circ\widehat{\mathcal{A}}'_1(\cdot)$. The trained adapters $\widehat{\mathcal{A}}'_1, \widehat{\mathcal{A}}'_2$ are then returned to the model owner to merge, combining as the final \textit{Plug-in} model $\mathcal{M}^* = \widehat{\mathcal{A}}'_2\circ\mathcal{E}\circ\widehat{\mathcal{A}}'_1(\cdot)$. Similarly, we also denote the directly fine-tuned $\mathcal{M}$ as $\mathcal{M}' = \mathcal{A}'_2\circ\mathcal{E}\circ\mathcal{A}'_1 (\cdot)$, where $\mathcal{A}'_1, \mathcal{A}'_2$ are the fine-tuned adapters with original $\mathcal{E}$.
% For simplification, we further denote $\widehat{\mathcal{M}}' = [\mathcal{A}’ , \mathcal{E}]$

Following the method of Lagrange multiplier, given a downstream dataset $\mathcal{D}$ that consists of training $D_{train}$ and testing $D_{test}$ sets, the objective of offsite tuning is to find $\mathcal{F}$ to minimize the test loss gap between $\mathcal{M}^*$ and $\mathcal{M}'$ while maximize the loss discrepancy between $\mathcal{M}^*$ and $\widehat{\mathcal{M}}'$:
\begin{align}
\nonumber
\min_{\mathcal{F}}  
&\underbrace{\mathcal{L}({\mathcal{M}^*, D_{test}) - \mathcal{L}(\mathcal{M}}', D_{test})}_{\text{Term (1): Plug-in performance preserving}} \\
& \underbrace{- \lambda \left (\mathcal{L}(\mathcal{M}^*, D_{test}) -\mathcal{L}(\widehat{\mathcal{M}}', D_{test}) \right )}_{\text{Term (2): Privacy protecting}}  
\\
\text{s.t. }  &
\nonumber\underbrace{\min_{\widehat{\mathcal{A}}'_1, \widehat{\mathcal{A}}'_2}\mathcal{L}(\widehat{\mathcal{M}}', D_{train}), \min_{\mathcal{A}'_1, \mathcal{A}'_2} \mathcal{L}(\mathcal{M}',D_{train})}_{\text{Conditions}}
\label{eq:offsite_tuning_obj}
\end{align}
\noindent where $\mathcal{L}(\cdot, \mathcal{D}) = \frac{1}{|\mathcal{D}|}\sum_{(x_i,y_i)\in \mathcal{D}}\ell(\cdot,x_i,y_i)$ is the average task loss of any given model on the dataset, $\lambda$ controls the privacy and utility trade-off of the offsite-tuning method, $\ell$ denotes the task loss function.

In this article, we explore the theoretical aspects of compressing models to achieve the goals of offsite tuning. In the following sections, we will first present our theoretical foundation and then propose methods to align with this framework.
 
\subsection{Forming Objectives of OT through the Perspective of Gradient}
This paper seeks an approach that meets Eq.~(1) of offsite tuning from the perspective of model weights and optimization. Specifically, we can regard the compression of weights as adding the noise $\delta$ to the weights, i.e., $\widehat{w} = w+\delta$. Given model with $n$ layers, $\mathcal{M} = \{m_1,m_2,\dots,m_n\}$, the loss of this model is:
\begin{equation}
\begin{aligned}
\ell(w, x_{j},y_{j}) =L(m_n(m_{n-1}(\cdots m_{1}(x_j,\\ w_{1})\cdots, w_{n-1}),w_n), y_{j}),
\end{aligned}
\end{equation}
\noindent where $W=\{w_1, w_2,\cdots,w_n\}$ denote the weight of each layer of $M$ in order. 
Considering the compressed weight $\widehat{W}=\{\hat{w}_1, \hat{w}_2,\cdots,\hat{w}_n\}$,
we only compress the weights of middle layers to generate the emulator, and the loss of emulator $\widehat{\mathcal{M}}$ is:
\begin{equation}
\begin{aligned}
\ell(\hat{w}, x_{j},y_{j}) =L(m_n(m_{n-1}(\cdots m_{1}(x_j,  \\ \hat{w}_{1})\cdots, \hat{w}_{n-1}),\hat{w}_n), y_j),\quad\\
 \hat{w}_i = \left\{
\begin{array}{ll}
w_i+\delta_i & \text{if } n_1 \le i \le n_2 \\
w_i & \text{otherwise}
\end{array}
\right. ,
\end{aligned}
\end{equation}
\noindent where $n_1$ and $n_2$ are the start and end layer index of the emulator.

Considering Term (1), the main problem is figuring out how much the gradients of the $i^{th}$ layer's weight $w_i$ would change while adding $\delta$ to the original model weights, which can be formed as:
\begin{equation}
\min \nabla_{w_i} \ell(\hat{w}) - \nabla_{w_i} \ell(w).
\end{equation}

In practice, it is almost impossible to estimate the gradient variations of the adapters through all layers of the emulator simultaneously due to memory limitations and computational cost. Based on the Chain Rule, we minimize the gradient variations of each layer, which ultimately presses the gradients of the adapters. Therefore, focusing on the gradient changes for a given weight $w_i$ that denoted as $\nabla_{w_i} \ell(w)$, while applying $\delta_i$, we have the following according to Taylor expansion:
\begin{equation}
\begin{aligned}
&\nabla_{w_i} \ell(\hat{w}) = \nabla_{w_i} \ell(w+\delta) \\
&\approx \nabla_{w_i} \ell(w)+\frac{\partial^2 \ell}{\partial w_i^2}\delta_i + \text{higher-order terms}, 
\end{aligned}
\end{equation}
where those higher-order terms can be omitted, and $\frac{\partial^2 \ell}{\partial w_i^2}$ is the partial hessian matrix of full model.

We then maximize the loss gap. To seek a tractable approximation of it, we need to expand $\ell(w)$ and $\ell(\hat{w})$ in the form of the \textit{total differential}. We remind readers of the total differential of the model.  

\begin{theorem}\label{theorem1}
The sum of the products of each partial derivative and the corresponding small change in the weight variable can estimate the increment in the loss function at a given point.
\end{theorem}

\noindent Theorem~\ref{theorem1} is the expression of the concept of total differential in weight factorization in neural networks. Through the definition of total differential and Theorem~\ref{theorem1}, we can build the connections between $\ell(w)$ and $\ell(\hat{w})$. For the differentiable function ${\ell}$ with the small variable $\delta$, the following equation is established:
\begin{equation}
\begin{aligned}
    % \ell(\hat{w})-\ell(w)=\sum_{i=n_1}^{n_2}\frac{\partial\ell}{\partial w_i}\odot\delta_{i}, \\
    \ell(\hat{w})=\ell(w) + \sum_{i=n_1}^{n_2}\frac{\partial\ell}{\partial w_i}\odot\delta_{i},
\end{aligned}
\label{eq6}
\end{equation}
where $\odot$ is the inner product. \cref{eq6} is obvious since $\hat{w}_i - w_i =\delta_i$ (see more details in \cref{app:diff_details}).

To achieve the goal of offsite tuning, a feasible solution is to ensure the gradient of the emulator is as close to raw as possible, but the divergence between the values of losses should be large:

\begin{equation}
\begin{aligned}
&\min_{\delta \in \Delta}  |\nabla_{w_i}  \ell(\hat{w}) - \nabla_{w_i} \ell(w)| \\&\quad\quad\quad\quad\quad\quad -\lambda(\ell(\hat{w}) - \ell(w)) , \\ 
&\approx  \underbrace{\sum^n ||\frac{\partial^2 \ell}{\partial w_i^2} \delta_i ||_1}_{\text{For Eq.~(1) Term (1)} } 
\underbrace{- \lambda(\sum^n\frac{\partial\ell}{\partial w_i}\odot\delta_{i})}_{\text{For Eq.~(1) Term (2)}}.
\end{aligned}
\label{eq:obj}
\end{equation}

\begin{figure*}[t]
\centering
\includegraphics[width=0.95\linewidth]{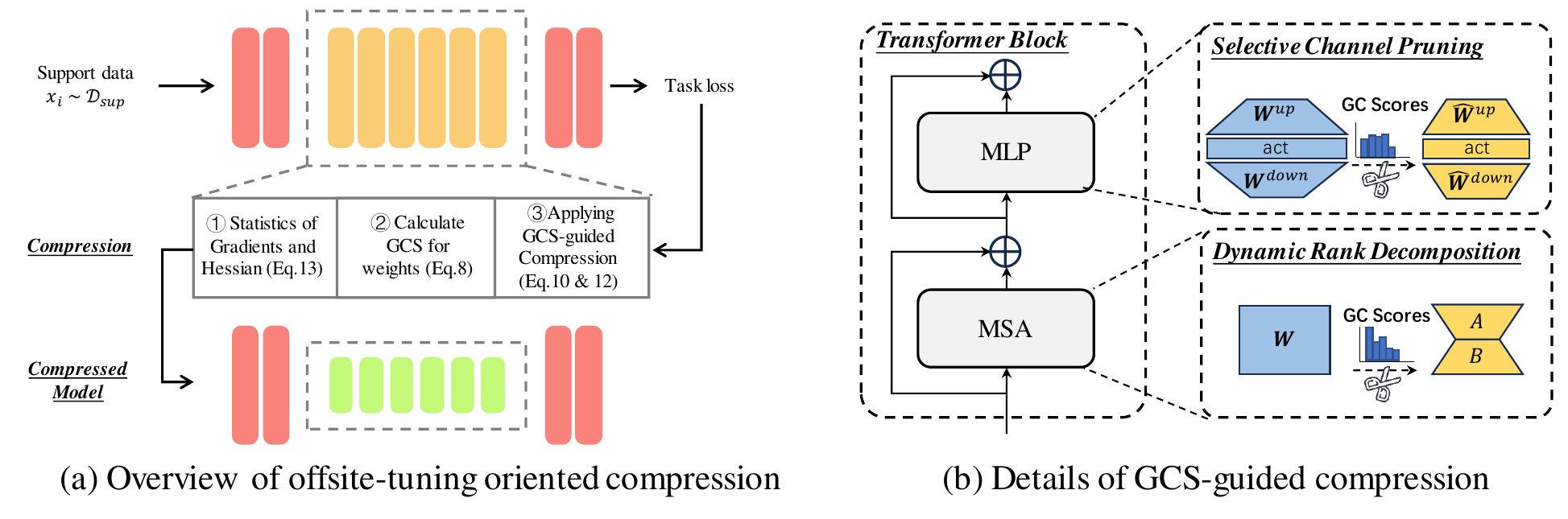}
\caption{Overview of our Gradient-preserving Compression Score (GCS) guided compression strategy.
}
\label{fig:method}
\end{figure*}

\subsection{Deriving OT method through Training-free  Compression}
\label{sec:Deriving OT method through Training-free  Compression}
This section demonstrates how to achieve \cref{eq:obj} empirically. We concentrate on the design of offsite-tuning for the Transformer architecture~\cite{transformers}, which is extensively used in LLMs~\cite{gpt2,gpt3,LLaMA}. A typical Transformer~\cite{transformers} layer is composed of two blocks: the Multi-Head Attention (MHA) and the Multilayer Perceptron (MLP). The MHAs facilitate interactions among tokens, while the MLPs further process information transformation within tokens. 
As the overall method shown in \cref{fig:method}~(a), we specifically employ Dynamic Rank Decomposition (DRD) on MHA and Selective Channel Pruning (SCP) on MLP as illustrated in \cref{fig:method}~(b). Both compression methods rely on \cref{eq:obj} to achieve optimal offsite-tuning. For each compressed weight $\widehat{w}_i$, we compute the Gradient-preserving Compression Score (GCS) for weight noise $\delta_i$ as follows:
\begin{equation}
{\rm GCS}(\delta_i) = \underbrace{||\frac{\partial^2 \ell}{\partial w_i^2} \delta_i ||_1}_{\text{Score Term (1)}} \underbrace{- \lambda\frac{\partial\ell}{\partial w_i}\odot\delta_{i}}_{\text{Score Term (2)}}.
\end{equation}
where the Score Term (1) and (2) correspond to the Eq.~(1) Term (1) and (2), respectively. 
Our goal is to determine $\delta_i$ by minimizing the scores while adhering to the given compression ratio. We set different trade-off factors $\lambda_{mha}$ and $\lambda_{mlp}$ for MHA and MLP, respectively.

\noindent \textbf{Dynamic Rank Decomposition.}
Rank compression is an effective matrix compression strategy, which reduces the size of a weight matrix by decomposing it into the product of two low-rank matrices. 
Considering the weight of a linear layer,
denoted as $\boldsymbol{W} \in \mathbb{R}^{d_{o}\times d_{i}}$, we utilize truncated Singular Value Decomposition (SVD)~\cite{eckart1936approximation} as:
\begin{equation}
\boldsymbol{W} = \boldsymbol{U} \boldsymbol{\Sigma} \boldsymbol{V}^T.
\label{eq:svd}
\end{equation}
For each weight in the MHA, given the compression ratio $\boldsymbol{r}_{mha}$, we generate the compressed weight as follow,
\begin{equation}
\begin{aligned}
&\widehat{\boldsymbol{W}} = \boldsymbol{B}\boldsymbol{A}, 
\boldsymbol{A} = \boldsymbol{V}^{T}_{s,:}, \boldsymbol{B} = \boldsymbol{U}_{:,s} \boldsymbol{\Sigma}_{s,s},\\
&\text{s.t.} \min_s {\rm GCS}(\widehat{\boldsymbol{W}}-\boldsymbol{W}), |s| = \frac{\boldsymbol{r}_{mha} d_i d_o}{d_i+d_o},
\end{aligned}
\label{eq:drd}
\end{equation}
where 
$\boldsymbol{A} \in \mathbb{R}^{|s| \times d_{i}}$, 
$\boldsymbol{B} \in \mathbb{R}^{d_{o} \times |s|}$, 
$s$ is the set of the rank index containing $|s|$ elements, 
$\boldsymbol{\Sigma}_{s,s}$ contains the selected singular values, and 
$\boldsymbol{U}_{:,s}$ and 
$\boldsymbol{V}_{s,:}$ are the corresponding singular vectors.
To ensure accurate scoring, we consistently select the top 5\% of ranks due to the constraint that $\delta_i$ should not be excessively large.

\noindent \textbf{Selective Channel Pruning.}
For expressive capability, the MLP's intermediate dimension is empirically set to be very high, significantly exceeding its input and output dimensions.
Formally, the formulation of MLP blocks is as follows:
\begin{equation}
f_{\rm MLP}(\boldsymbol{X})= \sigma(\boldsymbol{X}\boldsymbol{W}^{up})\boldsymbol{W}^{down},
\end{equation}
where $\boldsymbol{W}^{up} \in \mathbb{R}^{d_{h} \times d_{int}}$,  $\boldsymbol{W}^{down} \in \mathbb{R}^{d_{int} \times d_{h}} $, $\sigma$ is the activation function, $d_{h}$ and $d_{int}$ denote hidden dimension and intermediate dimension of LLM respectively, $d_{h} \ll d_{int}$. Similar to Eq.~\ref{eq:drd}, given compression ratio $\boldsymbol{r}_{mlp}$, we search for the optimal subset $s$ within the intermediate dimension to reduce parameters:
\begin{equation}
\begin{aligned}
&\widehat{f_{\rm MLP}}(\boldsymbol{X}) = \sigma(\boldsymbol{X}\boldsymbol{W}^{up}_{ :,s})\boldsymbol{W}^{down}_{s,:}, \\
& \text{s.t.} \min_s [{\rm GCS}(\boldsymbol{W}^{up}_{ :,s}-\boldsymbol{W}^{up})+ \\ 
&{\rm GCS}(\boldsymbol{W}^{down}_{s,:}-\boldsymbol{W}^{down})], |s| = \boldsymbol{r}_{mlp}d_{int}.
\end{aligned}
\label{eq:scp}
\end{equation}

\begin{algorithm}[t]
\caption{Gradient-preserving Compression}
\label{alg:ocot} 
\begin{algorithmic}[1] %[1] enables line numbers
\small
\State \textbf{Requires:} The LLM $\mathcal{M}$ with $n$ layers, the support dataset $\mathcal{D}_{sup}$, compression ratios $\boldsymbol{r}_{mha}$ and $\boldsymbol{r}_{mlp}$, trade-off factors $\lambda_{mha}$ and $\lambda_{mlp}$.
\ForAll{$i \in \{1, \ldots, |\mathcal{D}_{\text{sup}}|\}$}
    \State Compute the loss $\ell(w, x_i, y_i)$ 
    \State Perform backpropagation
    \State Accumulate gradients $\sum \frac{\partial \ell}{\partial w_i}$
    \State Accumulate approximate second-order gradients $\sum \frac{\partial^2 \ell}{\partial w_i^2}$ using Eq.~\eqref{eq:kfac}
\EndFor
\State Initialize Emulator $\widehat{\mathcal{E}}$ and Adapter $\widehat{\mathcal{A}}$
\ForAll{$w_i \in \widehat{\mathcal{E}}$}
    \If{$w_i$ belongs to MHA}
        \State $\widehat{w_i}={\rm DRD}(w_i,\boldsymbol{r}_{mha},\lambda_{mha})$ based on Eq.~\eqref{eq:drd}
    \ElsIf{$w_i$ belongs to MLP}
        \State $\widehat{w_i}={\rm SCP}(w_i,\boldsymbol{r}_{mlp},\lambda_{mlp})$ based on Eq.~\eqref{eq:scp}
    \EndIf
\EndFor
\State \textbf{return} $\widehat{\mathcal{E}}$,  $\widehat{\mathcal{A}}$
\end{algorithmic}
\end{algorithm}

\input{tables/slm}

\noindent \textbf{Approximation of Hessian Matrix.} 
In our score, we have to calculate the partial Hessian of each linear in the emulator.
Directly computing a linear's partial Hessian matrix demands substantial computational resources, particularly for LLMs. 
\citep{radhakrishnan2024mechanism} also highlights that the features extracted by a given neural network layer are proportional to the average gradient outer product with respect to the input of that layer.
Consequently, one feasible solution~\cite{kunstner2019limitations} is to estimate the Hessian using Fisher Information. Nonetheless, estimating the entire Hessian matrix remains exceptionally challenging because the full Hessian or Fisher matrix is a $P \times P$ matrix, where $P$ represents the number of parameters.

To mitigate this challenge, we utilize a Kronecker-factored approximation (KFAC) of individual weight matrices~\cite{daxberger2021laplace,ritter2018scalable,yang2023bayesian}, approximating the Fisher matrix through block structures within each linear layer. For the $i_{th}$ linear layer, we compute the block by denoting the input as $\boldsymbol{a}_{i-1}$ and output as $\boldsymbol{b}_{i}$. Then, the partial Hessian can be estimated by the following,
\begin{equation}
\begin{small}
 \frac{\partial^2 \ell}{\partial w_i^2} \approx \frac{1}{|\mathcal{D}|}\sum_{(x_i,y_i)\in \mathcal{D}}[(\boldsymbol{a}_{i-1}\boldsymbol{a}_{i-1}^T)\otimes(\boldsymbol{g}_{i}\boldsymbol{g}_{i}^T)],
 \label{eq:kfac}
 \end{small}
\end{equation}
where $\boldsymbol{g}_i = \frac{\partial\ell}{\partial \boldsymbol{b}_i}$ is the gradient of $\boldsymbol{b}_i$, $\otimes$ denotes Kronecker-factored multiplication.

\input{tables/llm}
\input{tables/llm2}

\noindent \textbf{Overall.} By leveraging the proposed scoring for the offsite-tuning objective, we can effectively assess the impact of weight noise on both loss and gradient metrics by utilizing the gradients and second-order derivatives of the original network. 
This method demonstrates greater efficiency compared to traditional approaches. 
Consequently, we introduce a support dataset $\mathcal{D}_{sup}$ to compute relevant values necessary for the score computation.
Notably, for an identical network, we only need to perform the statistical analysis once, thereby substantially reducing computational demands and time, as shown in \cref{sec:time}. Additionally, during the compression phase, we sequentially compress each weight in the emulator and determine the optimal results based on the calculated scores, significantly enhancing both efficiency and performance, as detailed in \cref{sec:efficient}. Our method is detailed in Algorithm~1 for clarity.

\section{Experiments}
\subsection{Experimental Setup}
\noindent \textbf{Datasets.} We validate our method across eight commensense datasets following previous works~\cite{scaleot}, including Multi-Choice QA: OpenBookQA~\cite{openbookqa}, PIQA~\cite{piqa},  SciQ~\cite{sciq}, RACE~\cite{race}; Closed-Book QA: ARC-Easy/Challenge~\cite{arc}; and Sentence Completion: HellaSwag~\cite{hellaswag}. We use \texttt{lm-eval-harness}~\footnote{{https://github.com/EleutherAI/lm-evaluation-harness}} to evaluate LLMs for a fair comparison. In order to preserve the privacy of the target datasets, we adhere \citet{crash} to generate the supporting dataset from the same tasks, resulting in a total of 1,500 samples across BoolQ~\cite{boolq}, TriviaQA~\cite{joshi2017triviaqa}, and CoPA~\cite{wang2019superglue}.

\noindent \textbf{Models.} We evaluate our method on large language models, including OPT-1.3B~\cite{opt}, OPT-6.7B~\cite{opt}, LLaMA-7B, and LLaMA-13B~\cite{LLaMA}.  In line with previous studies~\cite{OT}, we consistently select the first and last two layers as adapters, meaning that only about 10\% of the parameters are fine-tuned, as opposed to full fine-tuning. The remaining layers are then lossy compressed to generate the emulator. For compression, we set $[\boldsymbol{r}_{mha}, \boldsymbol{r}_{mlp}]$ to $[0.4, 0.7]$ for 1.5B model and $[0.5,0.8]$ for 7B models, resulting in emulators with 60\% and 70\% parameters respectively. We present implementation details about experiments in \cref{sec:details}.

\subsection{Main Results}
We conduct a comprehensive comparison of our method's offsite tuning performance against existing state-of-the-art (SoTA) methods. For training-based methods, our comparison includes OT~\cite{OT} and ScaleOT~\cite{scaleot}, while for training-free methods, we compare with Uniform LayerDrop (achieved by OT without knowledge distillation) and CRaSh~\cite{crash}.  Meanwhile, we present the results of full model zero-shot (ZS) and fine-tuning (FT),  which act as the baseline for this study.

\cref{tab:medlan}  presents the comparative results of offsite tuning on medium-sized models. All methods meet the offsite tuning conditions, meaning the plug-in's performance surpasses both the full model ZS and emulator FT. However, there remains a significant gap between the plug-in and full model FT performances, particularly for CRaSh and OT$^\dag$. Nonetheless, our method achieves promising plug-in results, with performance only slightly lower than that of full model FT. On the other hand, when compared with the SoTA lossless offsite-tuning method, ScaleOT, which utilizes dense post-training compression, our method offers superior privacy protection. Notably, our approach achieves the lowest emulator ZS performance, highlighting the efficacy of the proposed scoring mechanism that amplifies task loss. Overall, our method delivers the best privacy-utility trade-off, making it a promising alternative to existing training-based methods.

We subsequently validated the effectiveness of our method when scaling up to larger models, including OPT-6.7B, LLaMA-7B and LLaMA-13B, as shown in \cref{tab:biglan} and \cref{tab:llama13}. The results demonstrate that our proposed method exceeds previous training-free OT methods and achieves competitive plug-in performance compared to the existing post-training method, ScaleOT. 
Notably, we significantly outperform the training-free methods, OT and CRaSh, by an average of $4.1\%$  and $1.1\%$ on the LLaMA-7B model. Additionally, we consistently outperform OT on the LLaMA-13B model, further demonstrating our method's effectiveness on larger models.

One significant advantage of GradOT is its efficiency. As further detailed in Appendix~\ref{sec:time}, GradOT requires significantly less time to achieve results comparable to SoTA post-training methods.

\subsection{Analytical Studies}
This section presents key analytical studies while Appendix~\ref{app:more analytical analysis} provides additional analysis.

\input{tables/ablation}

\noindent \textbf{Impact of different components.}
The ablation study results, as presented in \cref{tab:ablation}, provide valuable insights into the impact of the different components proposed on performance across the OBQA, ARC-C, and SciQ datasets. We selected OBQA and SciQ as they represent the most and least challenging multi-choice QA datasets among the candidates, allowing us to assess GradOT's effectiveness across different difficulty levels. Additionally, ARC-C, being the most challenging Closed-Book QA dataset, helps validate our proposed components for such tasks. 

It is evident that the sole application of SCP improves plug-in performance but yields weak privacy protection, as indicated by the modest $\Delta = 1.3$. In contrast, using only DRD significantly enhances $\Delta$ compared to SCP alone, though it may lead to degradation in plug-in model performance. The combined use of both DRD and SCP achieves the highest compression ratio with the lowest $\boldsymbol{r}$ while also delivering the best $\Delta$. Furthermore, this combination results in plug-in performance comparable to that of the non-compressed model, confirming the effectiveness of both the DRD and SCP components.
Additionally, a comparison of the fourth-row results to the last two rows highlights the importance of the gradient-preserving compression score. Without this score, both plug-in performance and $\Delta$ experience a notable decline, underscoring the critical role of the proposed score in ensuring robust performance and privacy preservation. These findings collectively validate the efficacy of the individual components and the proposed score in achieving the desired balance between model compression, performance, and privacy protection.

\begin{figure}[t]
\centering
\includegraphics[width=.9\linewidth]{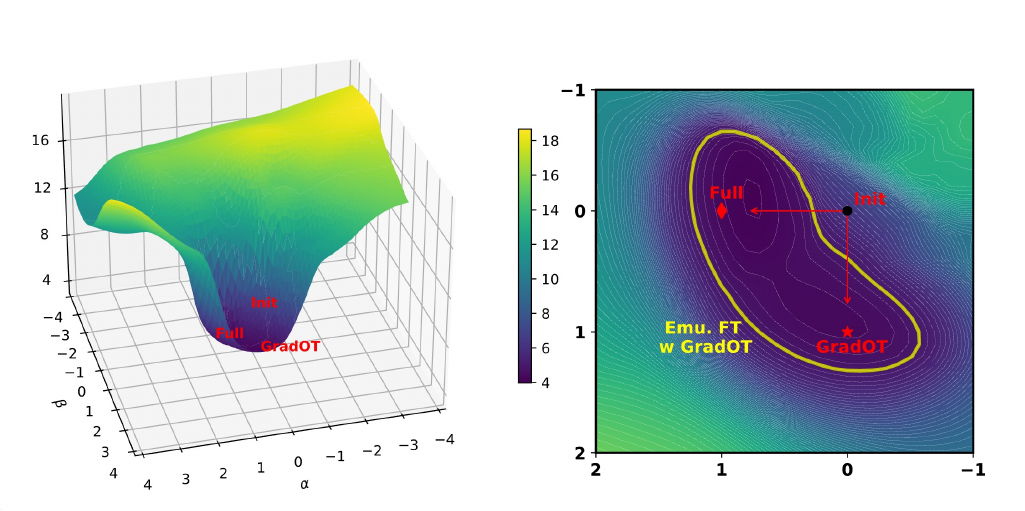}
\caption{Visualization of the loss landscape of the initialization weights and optima obtained through GradOT and full model fine-tuning. The loss level of Emu. FT with GradOT is also highlighted. 
}
\label{fig:loss_landscape}
\end{figure}

\noindent{\textbf{Loss Landscape Analysis.}} 
\cref{fig:loss_landscape} visualizes the loss landscape, illustrating that both the initialization weights and the optima obtained through GradOT and full model fine-tuning lie within the same basin and at the same level. This observation suggests that our gradient-preserving approach enables $\widehat{\mathcal{A}}'$ to converge to the same level as $\mathcal{A}'$, demonstrating the effectiveness of GradOT. Notably, the loss for Emu. FT with GradOT is positioned significantly further from both the fully fine-tuned and the GradOT Plug-in models, highlighting that GradOT sufficiently preserves privacy.

\begin{figure}[t]
\centering
\includegraphics[width=.85\linewidth]{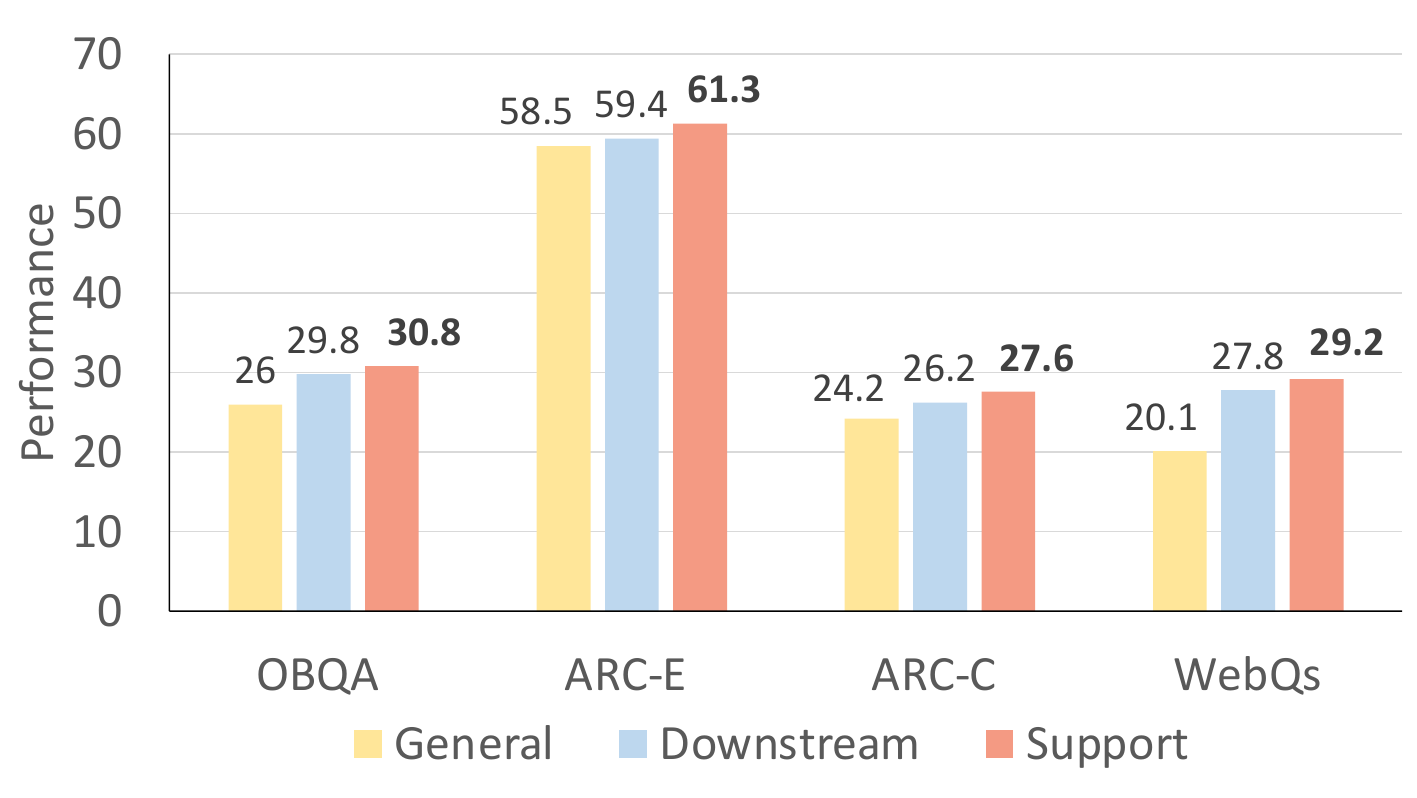}
\caption{Comparison of different data types for statistical analysis in terms of plug-in results of our method.
}
\label{fig:tab:datatype}
\end{figure}

\noindent \textbf{Impact of Data Type for Support Dataset.}
To protect the privacy of downstream data, we use a separate supporting dataset exclusively for statistical analysis to compute the first- and second-order derivatives, which are then used to calculate the Score. As shown in \cref{fig:tab:datatype}, we compare support datasets with general tasks using WikiText as the general dataset and the downstream dataset directly. The results obtained from the downstream dataset are significantly better than those from the general dataset. This is likely due to the differing contributions of various weights for general and downstream tasks. The support dataset yields the best results, even surpassing the downstream data. We hypothesize that this may be due to the supporting dataset containing multiple datasets, offering better generalization for downstream tasks.
In future work, we will explore how to select the support dataset to maximize generalizability.

\section{Discussions}
\noindent \textbf{Effect of Gradient Gap Term.}
One may notice that previous studies~\cite{kim2023squeezellm,osawa2023pipefisher,ren2023low} conduct importance estimation for model compression through the perspective of Fisher information, whose efficacy has been validated. Our methods also connect to these researches as we adapt Fisher information, but it is used to evaluate the gradient variations.  Our results indicate that preserving the gradient of adapters based on Fisher information is feasible. 

\noindent \textbf{Indicator for Other Methods.}
Besides the effective proposed GradOT, our gradient-preserving compression score also serves as an indicator for general offsite-tuning-oriented compression methods, such as feature-based low-rank compression~\cite{bolaco}. Possible applications in broader cases can be explored in future work. 

\section{Conclusion}
This paper demonstrates that the OT problem can be reformulated into a gradient-based form, from which the gradient-preserving compression score is derived. We introduce GradOT, a method that combines Dynamic Rank Decomposition and Selective Channel Pruning, both of which utilize the gradient-preserving compression score. Extensive experiments and analyses suggest that GradOT offers a promising solution for OT tasks.

\section*{Acknowledgements}
This work was supported by Ant Group Postdoctoral Programme.

\section*{Limitation}
One limitation of our approach is that the overall design of the proposed GradOT is based on the application to the transformer-based architecture, as described in \cref{sec:Deriving OT method through Training-free  Compression}. Therefore, GradOT may not be feasible for non-transformer-based models, e.g., Mamba. 
Notably, due to limited resources, we were unable to validate LLMs like the LLaMA3 70B. These larger models demonstrate enhanced language comprehension capabilities and, as a result, deliver superior performance.

\bibliography{custom}

\clearpage
\appendix
\input{sections/appendix}

\end{document}

%% file: tables/slm.tex
\begin{table*}[!t]
\centering
\small
%\setlength\tabcolsep{3.5 pt}
%\begin{adjustbox}{width=\textwidth}
\begin{tabular}{cc|cccccccc|cc}\toprule[1.5 pt]
\textbf{\textit{Method} }& \textbf{\textit{Setting}} & \textbf{OBQA} & \textbf{PIQA} & \textbf{ARC-E} & \textbf{ARC-C} & \textbf{Hella.} & \textbf{SciQ} & \textbf{WebQs} & \textbf{RACE} & \textbf{Avg.} & \textbf{$\Delta\uparrow$} \\\midrule
 \multicolumn{12}{l}{\textbf{\textit{Full Large Language Model}}} \\\midrule
\multicolumn{2}{c|}{Zero-shot (ZS)}& 23.4 & 71.6 & 56.9 & 23.5 & 41.5 & 84.4 & 4.6 & 34.2 & 42.5 & \multirow{2}{*}{-} \\
\multicolumn{2}{c|}{Fine-tuning (FT)}& 31.4 & 75.2 & 61.3 & 27.7 & 42.7 & 92.5 & 31.2 & 37.0 & 49.9 & \\\midrule\midrule
\multicolumn{12}{l}{\textbf{\textit{Post-Training Compression}}} \\\midrule
\multirow{3}{*}{\textbf{OT}} & Emu. ZS $\downarrow$ & 19.4 &68.7 &53.9 &21.5 &35.1 &80.9 &1.3 &33.0 & \underline{39.2}  & \multirow{3}{*}{\underline{2.4}} \\
 & Emu. FT $\downarrow$ & 24.8 & 71.6 & 58.1 & 26.1 & 37.0 & 92.2 & 24.3 & 38.6 &  \underline{46.6} & \\
 &  Plug-in $\uparrow$  & 29.0 & 74.5 & 59.4 & 27.8 & 43.3 & 92.9 & 26.2 & 38.9 & \underline{49.0} &  \\\midrule
\multirow{3}{*}{\textbf{ScaleOT}} & Emu. ZS $\downarrow$& 17.2 &63.1 &41.8 &19.5 &32.1 &59.9 &0.1 &27.2 &\textbf{32.6}  & \multirow{3}{*}{\textbf{3.7}} \\
& Emu. FT $\downarrow$ & 27.2 & 70.9 & 52.5 & 26.5 & 37.8 & 90.0 & 25.3 & 39.1 & \textbf{46.2} & \\
 &  Plug-in $\uparrow$  & 28.2 & 75.2 & 61.9 & 28.3 & 42.9 & 94.0 & 28.2 & 40.8 &  \textbf{49.9} & \\\midrule\midrule
\multicolumn{12}{l}{\textbf{\textit{Training-free Compression}}} \\\midrule
\multirow{3}{*}{\textbf{OT$^\dagger$}} & Emu. ZS $\downarrow$ &  13.8 & 58.4 & 34.9 & 19.0 & 27.0  & 49.8 & 0.0   & 22.7 & \underline{28.2}  & \multirow{3}{*}{\underline{3.3}} \\
 & Emu. FT $\downarrow$ &24.6 & 69.3 & 50.4 & 21.2 & 32.7  & 89.4 & 21.8  & 36.5 & \textbf{43.2}  & \\
 &  Plug-in $\uparrow$  &  26.4 & 72.7 & 58.3 & 23.0 & 41.2  & 90.8 & 21.4  & 37.9 & 46.5 &  \\\midrule
 \multirow{3}{*}{\textbf{CRaSh}} & Emu. ZS $\downarrow$ & 14.0 & 57.0 & 35.9 & 18.5 & 25.9  & 84.3 & 4.7   & 34.2&  34.3 & \multirow{3}{*}{\textbf{4.8}} \\
 & Emu. FT $\downarrow$ & 25.0 & 68.9 & 50.0 & 21.5 & 33.6  & 88.9 & 21.8  & 38.9 & \underline{43.6}  & \\
 &  Plug-in $\uparrow$  & 30.2 & 73.2 & 60.0 & 24.8 & 41.9  & 93.1 & 23.7  & 39.9 & \underline{48.4} &  \\\midrule
\multirow{3}{*}{\textbf{GradOT}} &  Emu. ZS $\downarrow$ & 12.2 & 55.8&33.8 &18.0 & 26.9& 49.4& 0.0&22.3 &  \textbf{27.3} & \multirow{3}{*}{\textbf{4.8}} \\
& Emu. FT $\downarrow$ & 27.0& 70.2 & 51.1& 23.8& 33.6 & 88.2 &  28.4& 38.7 & 45.0 & \\
 &  Plug-in $\uparrow$  & 30.8 & 73.6 & 61.3 &27.6 & 41.7  & 93.9 &29.2 & 40.6 & \textbf{49.8}&  \\
 \bottomrule[1.5pt]
\end{tabular}
%\end{adjustbox}
\caption{Comparative results of offsite-tuning with OPT-1.3B on eight datasets. $\Delta$ denotes the performance difference between Emulator (Emu.) fine-tuning and Plug-in. $\dagger$ denotes OT without knowledge distillation. Best in \textbf{bold} and second best in \underline{underline}.
}%.
\label{tab:medlan}
\end{table*}

%% file: tables/llm.tex
\begin{table*}[t]
\centering
\small
\begin{tabular}{r|cccccccc|c}\toprule[1.5pt]
\multicolumn{1}{c|}{\textbf{\textit{Setting}}} & \textbf{OBQA} & \textbf{PIQA} & \textbf{ARC-E} & \textbf{ARC-C} & \textbf{HellaSwag} & \textbf{SciQ} & \textbf{WebQs} & \textbf{RACE} & \textbf{Avg.} \\\midrule
\multicolumn{10}{c}{\textbf{OPT-6.7B}} \\\midrule
Full Model ZS & 27.6 & 76.2 & 65.5 & 30.6 & 50.5 & 90.1 & 8.8 & 38.2 & 48.4 \\
Emulator ZS & 19.6 & 68.8 & 53.6 & 24.7 & 38.1 &  83.0 & 2.4 & 31.6 & 40.2 \\
Emulator FT & 28.4 &73.6& 64.5 & 30.9 & 46.0 &  93.1 & 23.4 & 41.5 & 50.2 \\\midrule
\multicolumn{10}{c}{\small\textbf{\textit{Post-training Compression}}} \\ 
\textbf{ScaleOT} Plug-in & 33.2 & 78.1 & 70.1 & 35.3 & 52.2 & 95.7 & 33.9 & 45.3 & \textbf{55.5} \\ 
\multicolumn{10}{c}{\small\textbf{\textit{Training-free Compression}}} \\ 
\textbf{OT$^\dagger$} Plug-in & 33.8 & 77.7 & 66.8 & 33.9 & 52.1 & 91.9 & 23.9 & 44.1 & 53.0 \\
\textbf{CRaSh} Plug-in & 38.8 & 78.0 & 70.7 & 36.3 & 53.4 & 95.3 & 26.1 & 45.2 & 
\underline{55.3} \\
\rowcolor{mygray}\textbf{GradOT} Plug-in & 34.8 & 78.0 & 70.0 & 34.8& 52.4 &  95.3 & 32.7 & 45.1 & \textbf{55.4} \\\midrule\midrule
\multicolumn{10}{c}{\textbf{LLaMA-7B}} \\\midrule
Full Model ZS &28.2& 78.3 & 67.3 & 38.2 & 56.4 & 89.7 & 0.0 & 40.0 & 49.8 \\
Emulator ZS &  23.0& 72.1 & 52.8 & 30.0 & 45.2 & 72.6 & 0.0 & 32.3 & 41.0 \\
Emulator FT & 30.6 & 76.3& 63.3 & 32.6 & 49.4 & 91.8 & 29.6 & 43.2 & 52.1 \\\midrule
\multicolumn{10}{c}{\small\textbf{\textit{Post-training Compression}}} \\
\textbf{ScaleOT} Plug-in & 37.4 & 79.7 & 73.2 & 42.3 & 58.1 & 95.7 & 33.7  & 45.6 & \textbf{58.2} \\ 
\multicolumn{10}{c}{\small\textbf{\textit{Training-free Compression}}} \\
\textbf{OT$^\dagger$} Plug-in & 33.0 & 78.8 & 69.6 & 39.0 & 57.4 & 83.5 & 27.3 & 44.0 & 54.1 \\
\textbf{CRaSh} Plug-in & 34.6 & 80.0 & 71.3 & 41.8 & 58.4 & 95.1 & 29.8 & 45.6 & \underline{57.1} \\
\rowcolor{mygray}\textbf{GradOT} Plug-in & 33.8 & 80.5 & 72.8 & 41.7 & 57.9 &  95.4 &  38.3& 44.8 & \textbf{58.2} \\
\bottomrule[1.5pt]
\end{tabular}
\caption{Comparative results of offsite-tuning with large-size language models on $8$ question answering benchmarks.  }
\label{tab:biglan}
\end{table*}

%% file: tables/llm2.tex
\begin{table}[]
\centering
\small
\setlength\tabcolsep{4 pt}
\begin{tabular}{r|cccc}\toprule[1.5pt]
\multicolumn{1}{c}{\textbf{\textit{Setting}}} & \textbf{OBQA} & \textbf{ARC-E} & \textbf{ARC-C} & \textbf{WebQs} \\\midrule
Full Model ZS & 30.6 & 74.5 & 43.9 & 0.0 \\
Emulator ZS & 24.0 & 51.8 & 30.2 & 0.0 \\
Emulator FT & 32.6 & 66.7 & 39.2 & 37.9 \\\midrule
\textbf{OT$^\dagger$} Plug-in & 34.4 & 76.5 & 43.8 & 35.4 \\
\textbf{GradOT} Plug-in & \textbf{36.2} & \textbf{77.8} & \textbf{50.1} & \textbf{48.1}\\\bottomrule[1.5pt]
\end{tabular}
\caption{Comparative results of offsite-tuning on LLaMA-13B.  }
\label{tab:llama13}
\end{table}

%% file: tables/ablation.tex
\begin{table}[]
\centering
\small
\setlength\tabcolsep{3.3 pt}
\begin{tabular}{ccccccccc}\toprule
\multirow{2}{*}{DRD} & \multirow{2}{*}{SCP}&\multirow{2}{*}{$\boldsymbol{r}$}  & \multicolumn{2}{c}{OBQA} & \multicolumn{2}{c}{ARC-C} & \multicolumn{2}{c}{SciQ} \\
 &  & & P.in & $\Delta$ & P.in & $\Delta$& P.in & $\Delta$ \\\midrule
- & - & 1.0 &31.4 & 0.0 & 27.7 & 0.0 & 92.5  & 0.0  \\
- & $\checkmark$ & 0.8 & \textbf{34.0} & 2.0 & \textbf{29.5} & 1.7 & 93.3 & 1.3 \\
$\checkmark$  & - & 0.8&  31.0 &3.4 &  28.3 &3.4  & 93.1&2.6\\
$\checkmark$  & $\checkmark$ & 0.6& 30.8 & \textbf{3.8} & 27.6 &{3.8}&  \textbf{93.9}&\textbf{5.7} \\\hline
\textcolor{gray}{w/o S} & $\checkmark$ & 0.6& 30.0 & -0.2 &  28.9 & 4.0 & 93.8 & 4.2  \\
$\checkmark$  & \textcolor{gray}{w/o S} & 0.6&  29.8 & 1.0 & 25.8 & \textbf{4.4} &93.0 &3.5 \\\bottomrule
\end{tabular}
\caption{Ablation study. $\boldsymbol{r}$ denotes the overall parameter ratio of the emulator $\widehat{\mathcal{E}}$ maintained from the original $\mathcal{E}$, while \textcolor{gray}{w/o S}~represents the compression without our proposed score.}
\label{tab:ablation}
\end{table}

%% file: sections/appendix.tex
\section{More Details of Our Method}
\subsection{Total Differential Details }
\label{app:diff_details}
Given a function \( f(x_1, x_2, \dots, x_n) \) with variables \( x_1, x_2, \dots, x_n \), its total differential \( df \) can be expressed as:
\begin{equation}
df = \sum_{i=1}^{n} \frac{\partial f}{\partial x_i} dx_i, |dx_i|<\epsilon,
\end{equation}
where $\epsilon$ is a small value.
By considering the sample as a constant and treating $w$ as the input $x$, with $f = \ell(w)$ representing the loss calculation,  we can derive the total differential of $\ell(w)$ as follows:
\begin{equation}
    \partial\ell(w) = \sum_{i=n_1}^{n_2}\frac{\partial\ell}{\partial w_i}\odot dw_{i}.
\label{eq:lossgap}
\end{equation}
$dw_{i}$ can be considered as the $\delta_{i}$ between $w_{i}$ and $\hat{w}_{i}$.
Therefore, we have:
\begin{equation}
\begin{aligned}
    \ell(\hat{w})-\ell(w)=\sum_{i=n_1}^{n_2}\frac{\partial\ell}{\partial w_i}\odot\delta_{i},
\end{aligned}
\label{eq:6}
\end{equation}
where $\odot$ is the inner product. 

\subsection{Efficient Compression}
\label{sec:efficient}
To reduce time and computational cost during model compression, we calculate and rank the scores of each weight component in ascending order, then select the top $K$ components to satisfy the compression rate requirement. Specifically, in the Dynamic Rank Decomposition (DRD) method, a given weight component $w_{(k)}$ is expressed as the product of the $k^{th}$ eigenvalue and its corresponding eigenvector:
\begin{equation}
w_{k} = \boldsymbol{U}_{:,k} \boldsymbol{\Sigma}_{k,k}\boldsymbol{V}^{T}_{k,:}.
\end{equation}
Conversely, the Selective Channel Pruning (SCP) method represents the weight of the $k^{th}$ channel:
\begin{equation}
w_{k} = \{\boldsymbol{W}_{:,k}^{up}, \boldsymbol{W}_{k,:}^{down}\}
\end{equation}
We then compute the scores of these weight components ${\rm GCS}(w_{(k)})$ and, upon completion, select the components with the highest $K$ scores to aggregate them $\delta_i = \sum w_{(k)}$, where the selected index set is $s=\{k|{\rm Score}(w_{(k)}) < {\rm Score}(w_{(K)})$\}. This approach enables us to calculate the weight scores once, selecting those corresponding to the top $K$ scores in both DRD and SCP methods, thereby achieving compression. It is important to acknowledge that this method may introduce some errors during score calculation, as $|| \frac{\partial^2 \ell}{\partial w_i^2} \delta_i ||_1 \leq \sum || \frac{\partial^2 \ell}{\partial w_i^2} w_{(k)} ||_1$; however, it substantially enhances computational efficiency and eliminates the need for search techniques. This strategy significantly enhances the efficiency of the model compression.

\subsection{Implementation Details}
\label{sec:details}
For downstream training, we employ the AdamW Optimizer with a grid search of learning rates. 
We perform a learning rate tuning process on a grid of values and report the runs with the highest emulator performance, where \{2e-5, 5e-5, 1e-4, 2e-4, 3e-4\} for OPT-1.3B, OPT-6.7B, and LLaMA-7B. Due to the large range differences in our scoring calculations, we roughly searched for $\lambda$ within \{1e1, 1e2, 1e3, 1e4, 1e5\}, which implies that better parameters might exist if a more refined search is conducted. 
All experiments are conducted on a workstation with one A100 80G.

\subsection{Dataset Statistics}
For downstream tasks, the detailed dataset statistics can be referred to \cref{tab:dataset_description}. Meanwhile, we showcase the instructions formats of different datasets in \cref{tab:instruction_format_1,tab:instruction_format_2}.
\begin{table}[!h]
\centering
% \vspace{2.8mm}
\vspace{-5pt}
\setlength{\tabcolsep}{3pt}
\begin{tabular}
{lccr}
\toprule
Dataset  & \# Train & \# Test &Answer  \\\midrule
{\textbf{Downstream Tasks}}        &  &     & \\
\quad{OBQA}           & 5.0K &500      &Option \\
\quad{PIQA}          & 16.1K&1,830    &Option \\
\quad{ARC-E}          & 1.1K &2,376    &Option \\
\quad{ARC-C}         & 2.3K &1,172    &Option \\
\quad{HellaSwag}    & 39.9K&10,042   &Option \\
\quad{SciQ}    & 11.7k & 1,000  &Option \\
\quad{WebQs}    & 3.8k & 2,032  &Option \\
\quad{RACE}    & 62.4k & 3,498  &Option \\
\bottomrule
\end{tabular}
\vspace{-2pt}
\caption{The statistics of datasets for evaluation. \#~Train and \#~Test denote the number of training and test samples, respectively.}
\label{tab:dataset_description}
\vspace{-8pt}
\end{table}

\section{Additional Experiments and Analysis}
\label{app:more analytical analysis}
\subsection{Consumption of Compression Cost.}
\label{sec:time}
We present the time consumption associated with our proposed GradOT.
It is necessary to pre-compute the statistical information of the model on the support dataset. Although this process may be time-consuming, it worth noting that it only needs to be completed once for each model. Despite the significant memory demands inherent in gradient computation, our training-free approach allows for linear-by-linear computation, facilitating scalability to large models.
We report our results using a support dataset size of 1500 on a single 80G A100 GPU. 
Due to the memory limitation, for 7B models, we divided the emulator into 7 groups, with each group containing 4 transformer blocks for separate computation.
As shown in \cref{tab:time,tab:time2}, gradient calculation takes 8 minutes for OPT-1.3B and 97 minutes for LLaMA-7B. In contrast, model compression is considerably faster than the preparation stage, taking only 1 minute for OPT-1.3B and 8 minutes for LLaMA-7B. Compared with post-training-based methods like OT and ScaleOT, as shown in \cref{tab:time2}, which require tons of computation and time cost (e.g., OT and ScaleOT cost 60 and 5 hours on 1.5B models, respectively.), our method significantly reduces the time and computational costs involved in model compression.

\input{tables/compression}

\subsection{Sensitive Study}

\input{tables/sensitive}

The sensitivity study of $\lambda_{mha}$ and $\lambda_{mlp}$ in Table~\ref{tab:sensitive} examines their impact on the performance of the Emulator Fine-Tuned (Emu. FT) and Plug-in models on the OBQA dataset. 
It can be noticed that the Plug-in performance is not very sensitive to $\lambda_{mha}$. As $\lambda_{mha}$ increases, the Emu. FT performance consistently declines from 29.2 to 26.0, while the Plug-in model maintains relatively stable performance, decreasing only slightly from 31.6 to 30.2. This suggests that gradient-preserving compression score on multi-head attention (MHA) negatively affects the Emu. FT model and improving the model's privacy protection, the Plug-in model remains robust.

Similarly, increasing $\lambda_{mlp}$ results in a decline in Emu. FT performance drops from 29.6 to 27.0, while the Plug-in model also experiences a slight performance drop from 33.0 to 30.8 while $\lambda_{mlp}\leq1e3$.
Further increasing $\lambda_{mlp}$ to $1e3$ leads to a significantly declined performance for both Emu. FT and Plug-in. 

Notably, the Plug-in model consistently outperforms Emu. FT across all settings, reinforcing the effectiveness of the proposed approach in privacy protection and preserving Plug-in performance. 
The default hyperparameter settings for OPT-1.3B, \textbf{1e4} for $\lambda_{mha}$ and \textbf{1e2} for $\lambda_{mlp}$, strike a balance between Plug-in model performance and privacy protection. 

\subsection{Compression Ratio for Emulator.}

\begin{figure}[h]
\centering
\includegraphics[width=\linewidth]{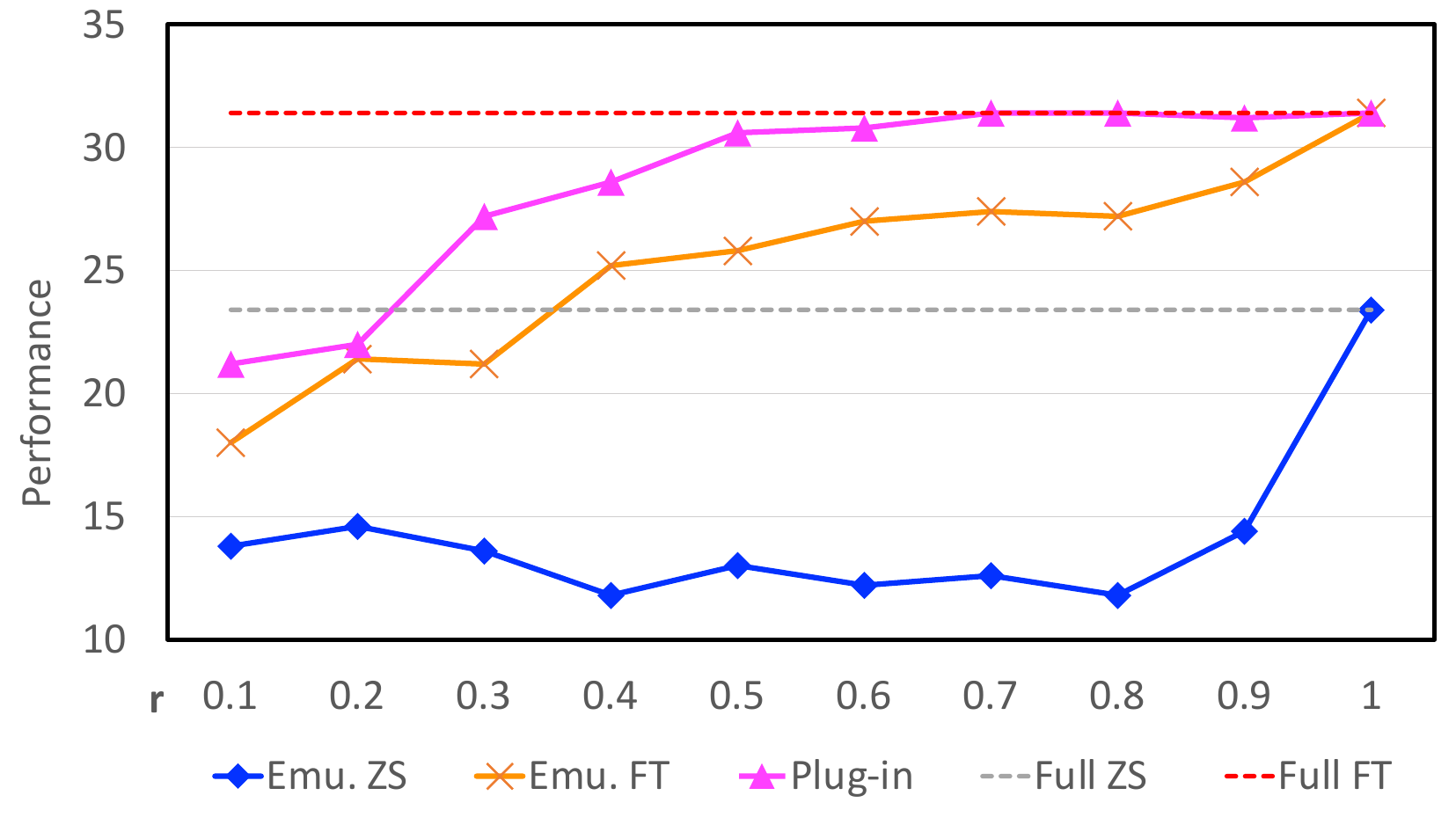}
\caption{Performance of applying GradOT to OPT-1.3B with increased $\boldsymbol{r}$ on OBQA dataset. 
}
\label{fig:compression}
\end{figure}
\cref{fig:compression} visualizes the results of applying GradOT to the OPT-1.3B model with varying values of $\boldsymbol{r}$. The constant gap between the plug-in results and the emulator's fine-tuned results for $\boldsymbol{r} <1$ suggests that GradOT can consistently preserve model privacy. Notably, when $\boldsymbol{r} \geq 0.5$, the plug-in model not only maintains privacy but also achieves performance comparable to, or even exceeding, that of the fully fine-tuned model. This highlights the effectiveness of GradOT in preserving privacy without sacrificing performance, making it a promising approach for OT with various compression ratios.

\subsection{Score Term (1) Approximates the Loss Discrepancy Between the Original and Noised Models.} 

\begin{figure}[h]
\centering
\includegraphics[width=1\linewidth]{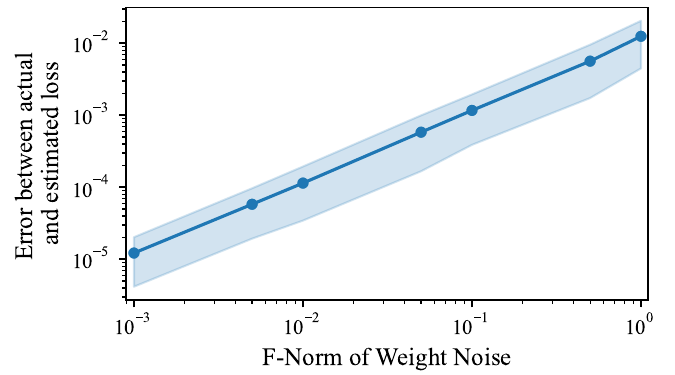}
\caption{Error between the actual loss and estimated loss with Eq.~\eqref{eq:lossgap} under different weight noise.
Both the horizontal and vertical axes are scaled by log.
}
\label{fig:apt}
\end{figure}

To evaluate the effectiveness of Score Term (1), we add random noise at varying scales into the middle-layer weights of the OPT-1.3B model and measure the error between the estimated values from Score Term (1) and the actual losses. Specifically, we progressively increase the noise magnitude, scaling its Frobenius norm from $0.001$ to $1$. As shown in \cref{fig:apt}, despite the increasing noise, the estimation error of Score Term (1) remains stable within $\pm 0.02$, demonstrating its robustness. These results validate the accuracy of Score Term (1) in approximating loss discrepancies and, consequently, support the efficacy of Score Term (2).

\input{tables/instruction_1}
\input{tables/instruction_2}

%% file: tables/compression.tex
\begin{table}[h]
\small
\centering
\begin{tabular}{ccc}\toprule
\textbf{Models} & Grad. Calculation & Model Compression \\\midrule
\textbf{OPT-1.3B} & 8 mins & 1 mins \\
\textbf{LLaMA-7B} & 97 mins & 8 mins \\\bottomrule
\end{tabular}
\caption{Time consumption of conducting  GradOT on different LLMs.}
\label{tab:time}
\end{table}

\begin{table}[h]
\small
\centering
\begin{tabular}{ccc}\toprule
\textbf{Methods} & \textbf{OPT-1.3B} & \textbf{LLaMA-7B} \\\midrule
\multicolumn{3}{l}{\small\textbf{\textit{Post-training Compression}}} \\
\quad OT & 60 hours & - \\
\quad ScaleOT & 5 hours & 57 hours \\
\multicolumn{3}{l}{\small\textbf{\textit{Training-free Compression}}} \\
\quad GradOT & 10 mins & 96 mins\\\bottomrule
\end{tabular}
\caption{Time consumption comparison on OPT-1.3B and LLaMA-7B.}
\label{tab:time2}
\end{table}

%% file: tables/sensitive.tex
\begin{table}[h]
\small
\centering
\setlength\tabcolsep{4 pt}
\begin{tabular}{ccccccc}
\cmidrule(lr){1-3} \cmidrule(lr){5-7}
\multirow{2}{*}{$\lambda_{mha}$} & \multicolumn{2}{c}{OBQA} &  & \multirow{2}{*}{$\lambda_{mlp}$} & \multicolumn{2}{c}{OBQA} \\
 & Emu. FT & Plug-in &  &  & Emu. FT & Plug-in \\ \cmidrule(lr){1-3} \cmidrule(lr){5-7}
 0 & 29.2 & 31.6 &  & 0 & 29.6 & 33.0 \\
1e3 & 28.6 & 31.2 &  & 1e1 & 29.2 & 30.8 \\ 
\textbf{1e4} & 27.0 & 30.8 &  & \textbf{1e2} &  27.0 & 30.8  \\ 
1e5 & 26.0 & 30.2 &  & 1e3 & 24.6 & 26.4 \\ 
\cmidrule(lr){1-3} \cmidrule(lr){5-7}
\end{tabular}
\caption{Sensitive study of $\lambda$ for OPT-1.3B. Default parameters are marked in \textbf{bold}.}
\label{tab:sensitive}
\end{table}

%% file: tables/instruction_1.tex
\begin{table*}[htbp]
\small
\centering
\begin{tabular}{p{15cm}l}
\toprule
\midrule
\rowcolor{gray!20} \textit{OpenBookQA} \\
\midrule
Poison causes harm to which of the following? \\
\textit{a Tree} \\
\midrule
As a car approaches you in the night \\
\textit{the headlights become more intense} \\
\midrule
\midrule
\rowcolor{gray!20} \textit{PIQA} \\
\midrule
Question: To create a makeshift ice pack, \\
Answer:\\
\textit{take a sponge and soak it in water. Put the sponge in a refrigerator and let it freeze. Once frozen, take it out and put it in a ziploc bag. You can now use it as an ice pack.} \\
\midrule
Question: What should I use as a stain on a wooden bowl I've just made. \\
Answer: \\
\textit{You should coat the wooden bowl with a butcher block oil \& finish per manufacturer directions.} \\
\midrule
\midrule
\rowcolor{gray!20} \textit{SciQ} \\
\midrule
One type of radioactivity is alpha emission. What is an alpha particle? What happens to an alpha particle after it is emitted from an unstable nucleus?.\\
Question: Alpha emission is a type of what?\\
Answer:\\
\textit{radioactivity} \\
\midrule
All radioactive decay is dangerous to living things, but alpha decay is the least dangerous.\\
Question: What is the least dangerous radioactive decay?\\
Answer:\\
\textit{alpha decay}\\
\midrule
\midrule
\rowcolor{gray!20} \textit{RACE} \\
\midrule
Article: Understanding the process of making career choices and managing your career is a basic life skill that everyone should understand.\\
Your career decisions have such a profound effect on all aspects of your life. It's important to have the knowledge and resources needed to make smart, informed decisions. Whether you are looking for a new job, aiming to take the next step at your current job or planning your retirement options, you are making career decisions. Using good resources and the guidance of a career counselor can help you to make those decisions well.\\
Many people mistakenly believe that choosing a career is a one-time event that happens some time in early adulthood. However, career management is actually a life-long process, and we continue to make consequential   career choices over the years. When people want to take action in their career, career management and job search are about so much more than writing a good resume. If you learn about and act on the following areas of career management, you'll be rewarded throughout your career.\\
Your interests, abilities, values, personal needs and realities should all be taken into account in any career decision making process. You spend countless hours at work, and it impacts your life in so many ways; it makes sense that you should be fully informed before making such profound decisions.\\
Do you know how many different career choices are available to you? Both The Dictionary of Occupational Titles (American) and The National Occupational Classification (Canadian) list well over 20,000 different job titles. So unless you've actively explored a variety of career options, there's a very good chance that there are great possibilities available to you, and you don't even realize they exist.\\
Match your understanding of yourself with your understanding of possible career options. Once you have developed a good understanding of yourself, you will be able to combine that self-knowledge with your career and labor market research to determine potential careers that are a great fit for you.\\
When you've made a well informed decision, then you're ready to make it happen. Making use of good career guidance and resources will help you to acquire the education, skills, and experience needed to get the job and learn and implement   effective job search strategies.\\
Time spent understanding your needs, researching your career options and developing outstanding job search skills, guided by great career resources, is a powerful investment in your future.\\
\\
Question: It can be inferred that  \_  .\\
\\
Answer:\\
\textit{career decision is misunderstood by many people because they don't take it as a life-long proces.} \\
\midrule
\bottomrule
\end{tabular}
\small
\caption{Instructions format of Multi-Choice QA task.}
\label{tab:instruction_format_1}
\end{table*}

%% file: tables/instruction_2.tex
\begin{table*}[htbp]
\small
\centering
\begin{tabular}{p{15cm}l}
\toprule
\midrule
\rowcolor{gray!20} \textit{ARC-E} \\
\midrule
Question: A chewable calcium carbonate tablet is a common treatment for stomach discomfort. Calcium carbonate is most likely used as this type of medicine because calcium carbonate\\
Answer:\\
\textit{neutralizes digestive acid.} \\
\midrule
Question: Which two body systems are directly involved in movement?\\
Answer:\\
\textit{muscular and skeletal} \\
\midrule
\midrule
\rowcolor{gray!20} \textit{ARC-C} \\
\midrule
Question: A fold observed in layers of sedimentary rock most likely resulted from the\\
Answer:\\
\textit{converging of crustal plates.}\\
\midrule
Question: As part of an experiment, an astronaut takes a scale to the Moon and weighs himself. The scale reads 31 pounds. If the astronaut has a mass of about 84 kilograms, which are the approximate weight and mass of the astronaut when standing on the Earth?\\
Answer:\\
\textit{186 pounds and 84 kilograms}\\
\midrule
\midrule
\rowcolor{gray!20} \textit{WebQS} \\
\midrule
Question: where is shoreview mn?\\
Answer:\\
\textit{Ramsey County}\\
\midrule
Question: what is the currency in the dominican republic called?\\
Answer:\\
\textit{Dominican peso}\\
\midrule
\midrule
\rowcolor{gray!20} \textit{HellaSwag} \\
\midrule
Getting a haircut: He then combs it and blow dries his hair after styling it with gel. He uses an electric clipper to groom the sideburns and the temples. He\\
\textit{also trims the back and sides of his head with the clippers.} \\
\midrule
Getting a haircut: The man in the center is demonstrating a hairstyle on the person wearing the blue shirt. The man in the blue shirt\\
\textit{sits on the chair next to the sink.} \\
\midrule
\bottomrule
\end{tabular}
\caption{Instructions format of Closed-Book QA and Sentece Completion task.}
\label{tab:instruction_format_2}
\end{table*}

%% file: main.bbl
\begin{thebibliography}{49}
\providecommand{\natexlab}[1]{#1}

\bibitem[{Bisk et~al.(2020)Bisk, Zellers, Gao, Choi et~al.}]{piqa}
Yonatan Bisk, Rowan Zellers, Jianfeng Gao, Yejin Choi, et~al. 2020.
\newblock Piqa: Reasoning about physical commonsense in natural language.
\newblock In \emph{AAAI}, volume~34, pages 7432--7439.

\bibitem[{Brown et~al.(2020)Brown, Mann, Ryder, Subbiah, Kaplan, Dhariwal, Neelakantan, Shyam, Sastry, Askell et~al.}]{gpt3}
Tom Brown, Benjamin Mann, Nick Ryder, Melanie Subbiah, Jared~D Kaplan, Prafulla Dhariwal, Arvind Neelakantan, Pranav Shyam, Girish Sastry, Amanda Askell, et~al. 2020.
\newblock Language models are few-shot learners.
\newblock \emph{Advances in Neural Information Processing Systems}, 33:1877--1901.

\bibitem[{Chua et~al.(2023)Chua, Yu, Zhao, and Lam}]{chua2023fedpeat}
Terence~Jie Chua, Wenhan Yu, Jun Zhao, and Kwok-Yan Lam. 2023.
\newblock Fedpeat: Convergence of federated learning, parameter-efficient fine tuning, and emulator assisted tuning for artificial intelligence foundation models with mobile edge computing.
\newblock \emph{arXiv preprint arXiv:2310.17491}.

\bibitem[{Clark et~al.(2019)Clark, Lee, Chang, Kwiatkowski, Collins, and Toutanova}]{boolq}
Christopher Clark, Kenton Lee, Ming-Wei Chang, Tom Kwiatkowski, Michael Collins, and Kristina Toutanova. 2019.
\newblock {B}ool{Q}: Exploring the surprising difficulty of natural yes/no questions.
\newblock In \emph{Association for Computational Linguistics}, pages 2924--2936.

\bibitem[{Clark et~al.(2018)Clark, Cowhey, Etzioni, Khot, Sabharwal, Schoenick, and Tafjord}]{arc}
Peter Clark, Isaac Cowhey, Oren Etzioni, Tushar Khot, Ashish Sabharwal, Carissa Schoenick, and Oyvind Tafjord. 2018.
\newblock Think you have solved question answering? try arc, the ai2 reasoning challenge.
\newblock \emph{arXiv preprint arXiv:1803.05457}.

\bibitem[{Daxberger et~al.(2021)Daxberger, Kristiadi, Immer, Eschenhagen, Bauer, and Hennig}]{daxberger2021laplace}
Erik Daxberger, Agustinus Kristiadi, Alexander Immer, Runa Eschenhagen, Matthias Bauer, and Philipp Hennig. 2021.
\newblock Laplace redux-effortless bayesian deep learning.
\newblock \emph{Advances in Neural Information Processing Systems}, 34:20089--20103.

\bibitem[{Devlin et~al.(2019)Devlin, Chang, Lee, and Toutanova}]{bert}
Jacob Devlin, Ming-Wei Chang, Kenton Lee, and Kristina Toutanova. 2019.
\newblock {BERT}: Pre-training of deep bidirectional transformers for language understanding.
\newblock In \emph{Proceedings of the 2019 Conference of the North {A}merican Chapter of the Association for Computational Linguistics: Human Language Technologies, Volume 1 (Long and Short Papers)}, pages 4171--4186.

\bibitem[{Dong et~al.(2024)Dong, Li, Dai, Zheng, Ma, Li, Xia, Xu, Wu, Chang, Sun, and Sui}]{dong2022survey}
Qingxiu Dong, Lei Li, Damai Dai, Ce~Zheng, Jingyuan Ma, Rui Li, Heming Xia, Jingjing Xu, Zhiyong Wu, Baobao Chang, Xu~Sun, and Zhifang Sui. 2024.
\newblock A survey on in-context learning.
\newblock In \emph{Conference on Empirical Methods in Natural Language Processing}, pages 1107--1128.

\bibitem[{Du et~al.(2022)Du, Qian, Liu, Ding, Qiu, Yang, and Tang}]{du2021glm}
Zhengxiao Du, Yujie Qian, Xiao Liu, Ming Ding, Jiezhong Qiu, Zhilin Yang, and Jie Tang. 2022.
\newblock {GLM:} general language model pretraining with autoregressive blank infilling.
\newblock In \emph{Association for Computational Linguistics}, pages 320--335.

\bibitem[{Eckart and Young(1936)}]{eckart1936approximation}
Carl Eckart and Gale Young. 1936.
\newblock The approximation of one matrix by another of lower rank.
\newblock \emph{Psychometrika}, 1(3):211--218.

\bibitem[{Hinton et~al.(2015)Hinton, Vinyals, and Dean}]{hinton2015distilling}
Geoffrey Hinton, Oriol Vinyals, and Jeff Dean. 2015.
\newblock Distilling the knowledge in a neural network.
\newblock \emph{arXiv preprint arXiv:1503.02531}.

\bibitem[{Hong et~al.(2024)Hong, Wang, Zhang, LI, Li, and Wang}]{dpopt}
Junyuan Hong, Jiachen~T. Wang, Chenhui Zhang, Zhangheng LI, Bo~Li, and Zhangyang Wang. 2024.
\newblock Dp-opt: Make large language model your privacy-preserving prompt engineer.
\newblock In \emph{International Conference on Learning Representations}.

\bibitem[{Ji et~al.(2024)Ji, Xiang, Li, Chen, Liu, Chen, and Zhang}]{bolaco}
Yixin Ji, Yang Xiang, Juntao Li, Wei Chen, Zhongyi Liu, Kehai Chen, and Min Zhang. 2024.
\newblock Feature-based low-rank compression of large language models via bayesian optimization.
\newblock \emph{In Findings of the Annual Meeting of the Association for Computational Linguistics}.

\bibitem[{Joshi et~al.(2017)Joshi, Choi, Weld, and Zettlemoyer}]{joshi2017triviaqa}
Mandar Joshi, Eunsol Choi, Daniel~S Weld, and Luke Zettlemoyer. 2017.
\newblock Triviaqa: A large scale distantly supervised challenge dataset for reading comprehension.
\newblock In \emph{Association for Computational Linguistics}.

\bibitem[{Kim et~al.(2024)Kim, Hooper, Gholami, Dong, Li, Shen, Mahoney, and Keutzer}]{kim2023squeezellm}
Sehoon Kim, Coleman Hooper, Amir Gholami, Zhen Dong, Xiuyu Li, Sheng Shen, Michael~W. Mahoney, and Kurt Keutzer. 2024.
\newblock Squeezellm: Dense-and-sparse quantization.
\newblock In \emph{International Conference on Machine Learning}.

\bibitem[{Kunstner et~al.(2019)Kunstner, Hennig, and Balles}]{kunstner2019limitations}
Frederik Kunstner, Philipp Hennig, and Lukas Balles. 2019.
\newblock Limitations of the empirical fisher approximation for natural gradient descent.
\newblock \emph{Advances in neural information processing systems}, 32.

\bibitem[{Lai et~al.(2017)Lai, Xie, Liu, Yang, and Hovy}]{race}
Guokun Lai, Qizhe Xie, Hanxiao Liu, Yiming Yang, and Eduard Hovy. 2017.
\newblock Race: Large-scale reading comprehension dataset from examinations.
\newblock In \emph{Conference on Empirical Methods in Natural Language Processing}, pages 785--794.

\bibitem[{Li et~al.(2020)Li, Ma, Guo, Xue, and Qiu}]{li2020bert}
Linyang Li, Ruotian Ma, Qipeng Guo, Xiangyang Xue, and Xipeng Qiu. 2020.
\newblock {BERT-ATTACK:} adversarial attack against {BERT} using {BERT}.
\newblock In \emph{Conference on Empirical Methods in Natural Language Processing}, pages 6193--6202.

\bibitem[{Liu et~al.(2024{\natexlab{a}})Liu, Li, Wu, and Lee}]{liu2024visual}
Haotian Liu, Chunyuan Li, Qingyang Wu, and Yong~Jae Lee. 2024{\natexlab{a}}.
\newblock Visual instruction tuning.
\newblock \emph{Advances in Neural Information Processing Systems}, 36.

\bibitem[{Liu et~al.(2024{\natexlab{b}})Liu, Liu, Tang, Yuan, Zhong, Chuang, Li, Chen, and Hu}]{liu2024lora}
Hongyi Liu, Zirui Liu, Ruixiang Tang, Jiayi Yuan, Shaochen Zhong, Yu-Neng Chuang, Li~Li, Rui Chen, and Xia Hu. 2024{\natexlab{b}}.
\newblock Lora-as-an-attack! piercing llm safety under the share-and-play scenario.
\newblock \emph{arXiv preprint arXiv:2403.00108}.

\bibitem[{Mihaylov et~al.(2018)Mihaylov, Clark, Khot, and Sabharwal}]{openbookqa}
Todor Mihaylov, Peter Clark, Tushar Khot, and Ashish Sabharwal. 2018.
\newblock Can a suit of armor conduct electricity? a new dataset for open book question answering.
\newblock In \emph{Conference on Empirical Methods in Natural Language Processing}.

\bibitem[{Nguyen et~al.(2021)Nguyen, Ding, Pathirana, Seneviratne, Li, and Poor}]{nguyen2021federated}
Dinh~C Nguyen, Ming Ding, Pubudu~N Pathirana, Aruna Seneviratne, Jun Li, and H~Vincent Poor. 2021.
\newblock Federated learning for internet of things: A comprehensive survey.
\newblock \emph{IEEE Communications Surveys \& Tutorials}, 23(3):1622--1658.

\bibitem[{Osawa et~al.(2023)Osawa, Li, and Hoefler}]{osawa2023pipefisher}
Kazuki Osawa, Shigang Li, and Torsten Hoefler. 2023.
\newblock Pipefisher: Efficient training of large language models using pipelining and fisher information matrices.
\newblock \emph{Proceedings of Machine Learning and Systems}, 5:708--727.

\bibitem[{Ouyang et~al.(2022)Ouyang, Wu, Jiang, Almeida, Wainwright, Mishkin, Zhang, Agarwal, Slama, Ray et~al.}]{ouyang2022training}
Long Ouyang, Jeffrey Wu, Xu~Jiang, Diogo Almeida, Carroll Wainwright, Pamela Mishkin, Chong Zhang, Sandhini Agarwal, Katarina Slama, Alex Ray, et~al. 2022.
\newblock Training language models to follow instructions with human feedback.
\newblock \emph{Advances in Neural Information Processing Systems}, 35:27730--27744.

\bibitem[{Radford et~al.(2019)Radford, Wu, Child, Luan, Amodei, Sutskever et~al.}]{gpt2}
Alec Radford, Jeffrey Wu, Rewon Child, David Luan, Dario Amodei, Ilya Sutskever, et~al. 2019.
\newblock Language models are unsupervised multitask learners.
\newblock \emph{OpenAI blog}, 1(8):9.

\bibitem[{Radhakrishnan et~al.(2024)Radhakrishnan, Beaglehole, Pandit, and Belkin}]{radhakrishnan2024mechanism}
Adityanarayanan Radhakrishnan, Daniel Beaglehole, Parthe Pandit, and Mikhail Belkin. 2024.
\newblock Mechanism for feature learning in neural networks and backpropagation-free machine learning models.
\newblock \emph{Science}, 383(6690):1461--1467.

\bibitem[{Ren and Zhu(2023)}]{ren2023low}
Siyu Ren and Kenny~Q Zhu. 2023.
\newblock Low-rank prune-and-factorize for language model compression.
\newblock \emph{arXiv preprint arXiv:2306.14152}.

\bibitem[{Ritter et~al.(2018)Ritter, Botev, and Barber}]{ritter2018scalable}
Hippolyt Ritter, Aleksandar Botev, and David Barber. 2018.
\newblock A scalable laplace approximation for neural networks.
\newblock In \emph{6th international conference on learning representations, ICLR 2018-conference track proceedings}, volume~6. International Conference on Representation Learning.

\bibitem[{Sajjad et~al.(2023)Sajjad, Dalvi, Durrani, and Nakov}]{layerdrop}
Hassan Sajjad, Fahim Dalvi, Nadir Durrani, and Preslav Nakov. 2023.
\newblock On the effect of dropping layers of pre-trained transformer models.
\newblock \emph{Computer Speech \& Language}, 77:101429.

\bibitem[{Sanh et~al.(2019)Sanh, Debut, Chaumond, and Wolf}]{sanh2019distilbert}
Victor Sanh, Lysandre Debut, Julien Chaumond, and Thomas Wolf. 2019.
\newblock Distilbert, a distilled version of bert: smaller, faster, cheaper and lighter.
\newblock \emph{arXiv preprint arXiv:1910.01108}.

\bibitem[{Singh et~al.(2024)Singh, Aditya, Madisetti, and Bahga}]{dptuning}
Tanmay Singh, Harshvardhan Aditya, Vijay~K. Madisetti, and Arshdeep Bahga. 2024.
\newblock Whispered tuning: Data privacy preservation in fine-tuning llms through differential privacy.
\newblock \emph{Journal of Software Engineering and Applications}, 17(1):1--22.

\bibitem[{Tan et~al.(2024)Tan, Yang, Wang, Nguyen, and Huang}]{tans}
Zhaorui Tan, Xi~Yang, Qiufeng Wang, Anh Nguyen, and Kaizhu Huang. 2024.
\newblock Interpret your decision: Logical reasoning regularization for generalization in visual classification.
\newblock In \emph{Advances in Neural Information Processing Systems}.

\bibitem[{Touvron et~al.(2023)Touvron, Martin, Stone, Albert, Almahairi, Babaei, Bashlykov, Batra, Bhargava, Bhosale et~al.}]{LLaMA}
Hugo Touvron, Louis Martin, Kevin Stone, Peter Albert, Amjad Almahairi, Yasmine Babaei, Nikolay Bashlykov, Soumya Batra, Prajjwal Bhargava, Shruti Bhosale, et~al. 2023.
\newblock Llama 2: Open foundation and fine-tuned chat models.
\newblock \emph{arXiv preprint arXiv:2307.09288}.

\bibitem[{Vaswani et~al.(2017)Vaswani, Shazeer, Parmar, Uszkoreit, Jones, Gomez, Kaiser, and Polosukhin}]{transformers}
Ashish Vaswani, Noam Shazeer, Niki Parmar, Jakob Uszkoreit, Llion Jones, Aidan~N Gomez, \L~ukasz Kaiser, and Illia Polosukhin. 2017.
\newblock Attention is all you need.
\newblock In \emph{Advances in Neural Information Processing Systems}, volume~30.

\bibitem[{Wang et~al.(2019)Wang, Pruksachatkun, Nangia, Singh, Michael, Hill, Levy, and Bowman}]{wang2019superglue}
Alex Wang, Yada Pruksachatkun, Nikita Nangia, Amanpreet Singh, Julian Michael, Felix Hill, Omer Levy, and Samuel Bowman. 2019.
\newblock Superglue: A stickier benchmark for general-purpose language understanding systems.
\newblock \emph{Advances in neural information processing systems}, 32.

\bibitem[{Wei et~al.(2022)Wei, Bosma, Zhao, Guu, Yu, Lester, Du, Dai, and Le}]{wei2021finetuned}
Jason Wei, Maarten Bosma, Vincent~Y. Zhao, Kelvin Guu, Adams~Wei Yu, Brian Lester, Nan Du, Andrew~M. Dai, and Quoc~V. Le. 2022.
\newblock Finetuned language models are zero-shot learners.
\newblock In \emph{International Conference on Learning Representations}.

\bibitem[{Welbl et~al.(2017)Welbl, Liu, and Gardner}]{sciq}
Johannes Welbl, Nelson~F. Liu, and Matt Gardner. 2017.
\newblock Crowdsourcing multiple choice science questions.
\newblock In \emph{EMNLP Workshop}, pages 94--106. Association for Computational Linguistics.

\bibitem[{Wies et~al.(2024)Wies, Levine, and Shashua}]{wies2024learnability}
Noam Wies, Yoav Levine, and Amnon Shashua. 2024.
\newblock The learnability of in-context learning.
\newblock \emph{Advances in Neural Information Processing Systems}, 36.

\bibitem[{Wortsman et~al.(2022)Wortsman, Ilharco, Kim, Li, Kornblith, Roelofs, Lopes, Hajishirzi, Farhadi, and Namkoong}]{wortsman2022robust}
Mitchell Wortsman, Gabriel Ilharco, Jong~Wook Kim, Mike Li, Simon Kornblith, Rebecca Roelofs, Raphael~Gontijo Lopes, Hannaneh Hajishirzi, Ali Farhadi, and Hongseok Namkoong. 2022.
\newblock Robust fine-tuning of zero-shot models.
\newblock In \emph{Proceedings of the IEEE/CVF Conference on Computer Vision and Pattern Recognition}, pages 7959--7971.

\bibitem[{Xiao et~al.(2023)Xiao, Lin, and Han}]{OT}
Guangxuan Xiao, Ji~Lin, and Song Han. 2023.
\newblock Offsite-tuning: Transfer learning without full model.
\newblock \emph{arXiv preprint arXiv:2302.04870}.

\bibitem[{Yang et~al.(2024)Yang, Robeyns, Wang, and Aitchison}]{yang2023bayesian}
Adam~X Yang, Maxime Robeyns, Xi~Wang, and Laurence Aitchison. 2024.
\newblock Bayesian low-rank adaptation for large language models.
\newblock In \emph{international conference on learning representations}.

\bibitem[{Yao et~al.(2024)Yao, Gao, Li, Zhao, Wang, Wang, and Zhu}]{yao-etal-2024-layer}
Kai Yao, Penglei Gao, Lichun Li, Yuan Zhao, Xiaofeng Wang, Wei Wang, and Jianke Zhu. 2024.
\newblock Layer-wise importance matters: Less memory for better performance in parameter-efficient fine-tuning of large language models.
\newblock In \emph{Findings of the Association for Computational Linguistics: EMNLP 2024}, pages 1977--1992.

\bibitem[{Yao et~al.(2025)Yao, Tan, Ye, Li, Zhao, Liu, Wang, and Zhu}]{scaleot}
Kai Yao, Zhaorui Tan, Tiandi Ye, Lichun Li, Yuan Zhao, Wenyan Liu, Wei Wang, and Jianke Zhu. 2025.
\newblock Scaleot: Privacy-utility-scalable offsite-tuning with dynamic layerreplace and selective rank compression.
\newblock In \emph{AAAI}.

\bibitem[{Ye et~al.(2024)Ye, Chen, Wang, Li, and Gao}]{bdatk}
Tiandi Ye, Cen Chen, Yinggui Wang, Xiang Li, and Ming Gao. 2024.
\newblock Bapfl: You can backdoor personalized federated learning.
\newblock \emph{Transactions on Knowledge Discovery from Data}, 18(7):166.

\bibitem[{Zellers et~al.(2019)Zellers, Holtzman, Bisk, Farhadi, and Choi}]{hellaswag}
Rowan Zellers, Ari Holtzman, Yonatan Bisk, Ali Farhadi, and Yejin Choi. 2019.
\newblock Hellaswag: Can a machine really finish your sentence?
\newblock In \emph{Association for Computational Linguistics}, pages 4791--4800.

\bibitem[{Zhang et~al.(2023{\natexlab{a}})Zhang, Ding, Qi, Zhu, Long, and Zhou}]{crash}
Kaiyan Zhang, Ning Ding, Biqing Qi, Xuekai Zhu, Xinwei Long, and Bowen Zhou. 2023{\natexlab{a}}.
\newblock {CR}a{S}h: Clustering, removing, and sharing enhance fine-tuning without full large language model.
\newblock In \emph{Conference on Empirical Methods in Natural Language Processing}, pages 9612--9637.

\bibitem[{Zhang et~al.(2023{\natexlab{b}})Zhang, Roller, Goyal, Artetxe, Chen, Chen, Dewan, Diab, Li, Lin et~al.}]{opt}
Susan Zhang, Stephen Roller, Naman Goyal, Mikel Artetxe, Moya Chen, Shuohui Chen, Christopher Dewan, Mona Diab, Xian Li, Xi~Victoria Lin, et~al. 2023{\natexlab{b}}.
\newblock Opt: Open pre-trained transformer language models.
\newblock \emph{arXiv preprint arXiv:2205.01068}.

\bibitem[{Zhou et~al.(2022)Zhou, Yang, Loy, and Liu}]{zhou2022conditional}
Kaiyang Zhou, Jingkang Yang, Chen~Change Loy, and Ziwei Liu. 2022.
\newblock Conditional prompt learning for vision-language models.
\newblock In \emph{Proceedings of the IEEE/CVF Conference on Computer Vision and Pattern Recognition}, pages 16816--16825.

\bibitem[{Zou et~al.(2023)Zou, Wang, Kolter, and Fredrikson}]{zou2023universal}
Andy Zou, Zifan Wang, J~Zico Kolter, and Matt Fredrikson. 2023.
\newblock Universal and transferable adversarial attacks on aligned language models.
\newblock \emph{arXiv preprint arXiv:2307.15043}.

\end{thebibliography}
